\documentclass[default,iicol]{sn-jnl}

\usepackage{graphicx}%
\usepackage{multirow}%
\usepackage{amsmath,amssymb,amsfonts}%
\usepackage{amsthm}%
\usepackage{mathrsfs}%
\usepackage[title]{appendix}%
\usepackage{xcolor}%
\usepackage{textcomp}%
\usepackage{manyfoot}%
\usepackage{booktabs}%
\usepackage{algorithm}%
\usepackage{algorithmicx}%
\usepackage{algpseudocode}%
\usepackage{listings}%

\theoremstyle{thmstyleone}%
\theoremstyle{thmstyletwo}%

\theoremstyle{thmstylethree}%

\begin{document}

\title[Article Title]{DHRL-FNMR: An Intelligent Multicast Routing Approach Based on Deep Hierarchical Reinforcement Learning in SDN}


\author*[1]{\fnm{Miao} \sur{Ye}} \email{yemiao@guet.edu.cn}

\author[1]{\fnm{Chenwei} \sur{Zhao}} \email{zhaochenwei96@foxmail.com}
\author[2]{\fnm{Xingsi} \sur{Xue}}
\author[1]{\fnm{Jinqiang} \sur{Li}} 
\author[1]{\fnm{Hongwen} \sur{Hu}} 
\author[1]{\fnm{Yejin} \sur{Yang}} 
\author[1]{\fnm{Qiuxiang} \sur{Jiang}} 

\affil[1]{\orgdiv{School of Information and Communication}, \orgname{Guilin University of Electronic Technology}, \orgaddress{\city{Guilin}, \postcode{541004}, \state{Guangxi}, \country{China}}}

\affil[2]{\orgdiv{Fujian Provincial Key Laboratory of Big Data Mining and Applications}, \orgname{Fujian University of Technology}, \orgaddress{\city{Fuzhou}, \postcode{350118}, \state{Fujian}, \country{China}}}

\abstract{The optimal multicast tree problem in the Software-Defined Networking (SDN) multicast routing is an NP-hard combinatorial optimization problem. Although existing SDN intelligent solution methods, which are based on deep reinforcement learning, can dynamically adapt to complex network link state changes, these methods are plagued by problems such as redundant branches, large action space, and slow agent convergence. In this paper, an SDN intelligent multicast routing algorithm based on deep hierarchical reinforcement learning is proposed to circumvent the aforementioned problems. First, the multicast tree construction problem is decomposed into two sub-problems: the fork node selection problem and the construction of the optimal path from the fork node to the destination node. Second, based on the information characteristics of SDN global network perception, the multicast tree state matrix, link bandwidth matrix, link delay matrix, link packet loss rate matrix, and sub-goal matrix are designed as the state space of intrinsic and meta controllers. Then, in order to mitigate the excessive action space, our approach constructs different action spaces at the upper and lower levels. The meta-controller generates an action space using network nodes to select the fork node, and the intrinsic controller uses the adjacent edges of the current node as its action space, thus implementing four different action selection strategies in the construction of the multicast tree. To facilitate the intelligent agent in constructing the optimal multicast tree with greater speed, we developed alternative reward strategies that distinguish between single-step node actions and multi-step actions towards multiple destination nodes. Finally, a series of experiments and their results show that compared with reinforcement learning and other algorithms, the designed algorithm can search for the optimal multicast trees more efficiently, converge faster and eliminate redundant branches, and the constructed multicast trees have better performance in terms of bandwidth, delay, and packet loss rate.}


\maketitle
\section{Introduction}\label{sec1}
The increasing prevalence of "one-to-many" multicast data transmission in network applications is a result of the rapid development of network infrastructure. By transmitting multicast data only once along the same path instead of multiple times, it is possible to save network traffic and reduce network congestion. To achieve this, it is necessary to find a suitable fork node for constructing an optimal multicast tree and transmitting multicast data along that tree.

The problem of constructing the optimal multicast tree can be modeled as a classical discrete optimization problem in mathematical graph theory - the Steiner tree problem. However, the dynamic changes in network link states at different times pose challenges to the construction of the optimal multicast tree. Therefore, timely and convenient collection and acquisition of network state information, as well as the design of algorithms for constructing the optimal multicast tree, are indispensable for multicast routing of data. Compared with traditional network measurement methods for obtaining network state information, Software-Defined Networking (SDN) technology, which adopts a structure with a separation between the data plane and the control plane, can effectively solve the problems of limited network state information, poor adaptability to dynamic and complex network changes, and inflexible data forwarding. By communicating with the data plane through the southbound interface of the SDN controller, centralized control and easy configuration and management of the network can be achieved, and data plane SDN switches can be used for state collection and forwarding routing control by means of flow table distribution. The centralized control programming method of the SDN controller provides a possibility for the implementation of multicast routing algorithms. Therefore, this article mainly focuses on the design and implementation of SDN multicast routing algorithms for dynamically changing networks under the SDN architecture.

The Steiner tree problem, which corresponds to the multicast tree problem, is an NP-complete problem \cite{R1}. Classical algorithms like the KMB algorithm (proposed by Kou, Markousky, and Berman) \cite{R2}, the MPH algorithm (Minimum Cost Path Heuristic) \cite{R3}, and the ADH algorithm can approximate the Steiner tree. However, they fail to take into account global dynamic changes in the network and are not suitable for solving large-scale problems. Intelligent optimization algorithms such as the Ant Colony Optimization (ACO) \cite{R4} and Genetic linewidthorithm (GA) \cite{R5} suffer from time-consuming and convergence issues to global optimization. Therefore, alternative methods are required to solve the multicast tree problem effectively.

Reinforcement learning (RL) methods aim to maximize reward through agent-environment interaction. Offline learning can train the agent without online computation, reducing computational time costs. RL has improved routing path construction \cite{R6, R7}, but current multicast research RL methods consider only single factors, such as hop count, and rarely consider multiple factors. In our previous work, the DRL-M4MR algorithm \cite{R8} was proposed to address this issue, but uses all links as the action space, leading to slow convergence due to negative sample selection. This makes it easy for the agent to select negative samples during the early stages of learning, leading to slow convergence speed. Additionally, redundant branches can occur during multicast tree construction.

To address the issues previously mentioned, this paper proposes an SDN intelligent multicast routing algorithm called HDRL-FNMR (\textbf{F}ork \textbf{N}ode find path to construct \textbf{M}ulticast \textbf{R}outing based on \textbf{D}eep \textbf{H}ierarchical \textbf{R}einforcement \textbf{L}earning) based on deep hierarchical reinforcement learning (DHRL). We model the process of constructing multicast trees as a semi-Markov decision process (SMDP). By performing multiple actions in a single execution, the agent can reach states in further time steps in SMDP \cite{R9}. Therefore, the process of constructing multicast trees can be divided into two sub-problems. The first is the upper-level sub-problem of selecting a fork node, while the second is the lower-level sub-problem of constructing the optimal path from the fork node to the destination node. Once the intrinsic and extrinsic controllers have built the optimal unicast paths from the fork node to the destination node multiple times, the agent can construct the optimal multicast tree.

This paper utilizes SDN's global network awareness feature to enable the agent to perceive global network information. A multicast tree state matrix, link bandwidth matrix, link delay matrix, link packet loss rate matrix, and sub-goal matrix are designed as the state space for the intrinsic and extrinsic controllers. To address the issue of a large action space for the agent, different action spaces are designed for the upper and lower levels. Specifically, the meta-controller considers all nodes in the network as the action space, while the intrinsic controller considers the neighboring edges of the current node as the action space. Four action selection strategies are designed to construct paths, and five reward strategies are designed to guide the agent in constructing the optimal multicast tree.

The points of innovation in this paper are as follows:
\begin{enumerate}[1.]
\item Based on our literature research, it is proposed for the first time to use the DHRL method to solve the multicast tree problem. An SDN multicast routing mechanism based on DHRL is designed, which divides the multicast tree construction process into two sub-problems and assigns them to the meta-controller and the intrinsic controller of the intelligent agent. The meta-controller solves the problem of selecting fork nodes, while the intrinsic controller solves the problem of constructing the optimal path from the fork node to the destination node.
\item In response to the characteristics of the SDN multicast problem, a state space is designed for the meta-controller consisting of the multicast tree state matrix, the link bandwidth matrix, the link delay matrix, and the link packet loss rate matrix. For the intrinsic controller, a state space consisting of the sub-goal matrix representing the position of the fork node and the above four matrices is designed, which enables the intelligent agent to better perceive changes in network status information and changes in the multicast tree construction process.
\item In response to the background characteristics of the two sub-problems of splitting, corresponding efficient action policies are designed for the DHRL at different levels, which greatly reduce the production of invalid actions and make the intelligent agent search more efficient. The action space of the designed meta-controller is the set of all nodes in the network topology, and each action selects one of them as a fork node. The size of the action space for the designed intrinsic controller is the maximum degree of the network topology, and the output action space tensor of the neural network corresponds one-to-one with neighboring edges of the current node.
\item Corresponding reward mechanisms are designed for the different action policies at different levels of DHRL, allowing the intelligent agent to receive more effective feedback and learn more efficiently through interaction with the environment. The reward function of the meta-controller is designed with rewards for reaching a single destination node, rewards for reaching all destination nodes, and punishments for invalid fork nodes, allowing the meta-controller to select the optimal sub-goal according to the current state and pass it down to the lower layer. The intrinsic controller's reward function is designed with rewards for single-step decisions, rewards for completing the construction path, penalties for invalid actions, and penalties for loops, allowing the intrinsic controller to quickly construct the optimal path from the upper fork node to the destination node.
\end{enumerate}

The rest of the paper is organized as follows: Section \ref{sec2} reviews related work on multicast routing. Section \ref{sec3} presents the problem description and decomposition for SDN multicast routing. Section \ref{sec4} describes the intelligent SDN multicast routing optimization architecture and modeling. Section \ref{sec5} details the design of the DHRL-FNMR algorithm's agent and the implementation details for multicast routing. Section \ref{sec6} presents the experimental results. Section \ref{sec7} concludes the paper with recommendations for future work.

\section{Related work}\label{sec2}

This section introduces the relevant methods for solving the optimal multicast tree and Steiner tree problems.

The classic solution methods, including KMB algorithm, MPH algorithm, and ADH (Average Distance Heuristic) algorithm are summarized below. The KMB algorithm \cite{R10} constructs a complete graph and utilizes the minimum spanning tree algorithm to obtain a path tree from the source node to the destination node in the graph.
Subsequently, the algorithm replaces the edges of the path tree in the complete graph with the shortest path algorithm to obtain an approximate optimal Steiner tree. Its time complexity is $O\left(mn^2\right)$. The MPH algorithm has a time complexity of $O\left(mn^2\right)$.  This algorithm starts with a multicast tree set that only includes the source node. Subsequently, it uses the shortest path algorithm to add the shortest path to the multicast tree set until all the destination nodes are included. The ADH algorithm, with a time complexity of $O\left(n^3\right)$ \cite{R11}, initially takes a multicast tree set that contains both the source node and all destination nodes. Subsequently, the algorithm selects the node with the minimum distance to connect any two disconnected trees until all trees are joined. He et al. proposed a heuristic algorithm for constructing reliable multicast trees with multiple sources, taking into account bandwidth, delay, and reliability constraints to ensure quality of service for multicast in virtual network functions. They utilized the Floyd algorithm and Prim algorithm \cite{R14} to construct the multicast trees. Shi et al. improved the shortest path in the KMB algorithm by selecting the path with fewer additional nodes (switches) between the shortest path and the second shortest path, reducing the number of flow entries in SDN switches \cite{R15}. Li et al. abstracted the problem into a Betweenness Centrality to Bandwidth ratio Tree (BCBT) problem \cite{R16}. They used the shortest path tree to generate multicast routing under low traffic and a heuristic greedy algorithm to solve the BCBT optimization problem under link congestion. Bijur et al. used the Dijkstra shortest path algorithm to construct multicast paths based on receiving devices and network capacity and handled dynamic device join and leave to reduce tree changes \cite{R17}. Hwang et al. proposed a Wireless Shortest Path Heuristic (W-SPH) mechanism for constructing multicast trees in wireless networks \cite{R18}. Sun et al. proposed a multicast routing algorithm for QoS assurance in network function virtualization for mobile edge computing, using the shortest path tree algorithm to construct multicast trees \cite{R19}. Sufian et al. proposed an energy- and velocity-based multicast protocol in Ad-Hoc Networks. Destination nodes select the route with the highest weight based on the route-request packet \cite{R20}. These algorithms have high computational complexity, slow convergence speed, and are difficult to obtain the optimal solution as the problem size increases.

Kotachi et al. used integer linear programming to minimize the total number of flow table entries \cite{R21}. Latif et al. used integer linear programming to improve the utilization of the fabric in CLOS networks \cite{R22}. Touihri et al. employed mixed integer linear programming to improve the minimum residual bandwidth and minimize the number of links in CamCube data center networks based on SDN \cite{R23}. Risso et al. proposed a flow-based mixed integer programming formulation for modeling and solving the problem of quality of service multicast tree \cite{R24}. However, the broad applicability of these methods may be constrained due to the extensive computational complexity of linear programming and the specific constraints designed for certain scenarios.

Intelligent optimization algorithms usually employ iterative procedures to solve optimization problems. Hassan et al. utilized the Ant Colony Optimization (ACO) algorithm to optimize routing problems. However, the computational time required was high \cite{R25}. Zhang et al. employed Leaf Crossover Genetic linewidthorithm (LCGA) with a leaf crossover mechanism, which improved the algorithm's performance, but execution time was long \cite{R5}. Shakya et al. utilized Bi-Velocity Particle Swarm Optimization (BVDPSO), a double-speed strategy, to solve multicast routing problems, resulting in faster convergence rates \cite{R26}. Sahoo et al. combined Particle Swarm Optimization (PSO) and Bacterial Foraging linewidthorithm (BFA) to construct QoS multicast routing \cite{R27}, but the iteration time required by the algorithm was large. To construct multicast trees with delay and bandwidth constraints, Murugeswari et al. proposed a PSO algorithm combined with GA \cite{R28}. Zhang et al. proposed a hybrid ACO and Cloud Model (CM) algorithm to improve search efficiency and avoid getting trapped in local optima \cite{R29}. Liu et al. employed a non-coding genetic algorithm to solve application-layer multicast tasks. This algorithm directly represented the genotype as a tree-like phenotype and recombined and mutated it to adapt to the tree-like genotype \cite{R30}. Despite their effectiveness, these algorithms encounter a common challenge when multiple multicast trees need to be built simultaneously: the computational power required is significant, making it impractical to compute routing strategies for each new flow in the SDN controller \cite{R31}.

Reinforcement learning approaches enable the construction of optimal paths through the interaction between an agent and the environment. Numerous studies have explored the use of reinforcement learning methods for unicast routing. Hendriks et al. used Q-learning algorithm to construct QoS-compliant unicast paths in ad hoc wireless networks. Each node selected a neighbor based on their Q-values for learning to forward a Q-info packet \cite{R32}. Casas-Velasco et al. proposed using Q-learning algorithm to tackle the unicast routing challenge in SDN. In this algorithm, the state space is designated as the source-destination pair while neighbor nodes are used as the action space. The reward is determined by aspects such as remaining bandwidth, delay, and packet loss rate \cite{R6}. An improvement was later suggested to use a Deep Q network (DQN) agent in real-time to select a path from k-paths for building unicast paths \cite{R33}. Guo et al. designed a QoS-aware secure routing protocol using the Deep deterministic policy gradient (DDPG) algorithm to construct unicast paths in SDN-IoT. The state space contains information such as the current packet arrival frequency, flow table occupancy rate, and occupancy rate linking switches and SDN controller. Meanwhile, the action space includes all switches within the network topology, and an action is chosen based on the probability of each action \cite{R34}. Yu et al. proposed a DDPG-based algorithm to develop unicast routes in SDN. In this algorithm, the state space is established as the current traffic matrix, while the action space is the weight of each link. The shortest path algorithm is involved in determining the route, and rewards are calculated reliant on delay, throughput, hop count, and other metrics \cite{R35}. Ning et al. and Kim et al. reassigned various values to links in the network topology and utilized conventional shortest path algorithms to determine routing paths \cite{R36,R37}. Nonetheless, these techniques are specific to unicast situations, and their designated state space, action space, and reward mechanisms cannot be directly applied to resolving multicast routing problems.

The FORMS algorithm was proposed by Forster et al. to construct multicast trees in wireless sensor networks (WSNs) using Q-learning. Initially, each node announces its presence by broadcasting, and then exchanges rewards using DATA packets \cite{R38}. However, the size of the state space and action space in Q-learning is a drawback. Heo et al. used DQN to solve the multicast routing problem in SDN, but only considered optimizing the length of the multicast tree \cite{R39}. Chae et al. suggested using the Advantage Actor-Critic (A2C) algorithm to handle multicast issues with network topologies that frequently change \cite{R40}. However, they also solely focused on optimizing the length of the multicast tree without considering other network information. Tran et al. proposed the DQMR protocol \cite{R41}. The protocol constructs unicast paths from the source node to the destination node multiple times using DQN in a mobile ad hoc network (MANET), and then combines them to form a multicast tree, thereby ensuring QoS for data transmission. However, this approach necessitates reconstructing the path from the origin node when building the multicast tree.

\section{SDN multicast problem}\label{sec3}
\subsection{steiner tree problem}\label{sec3.1}
The aim of this paper is to address the issue of multicast data transmission, which entails transmitting data from one source node to multiple destination nodes. The topology that connects a source node as the root node and multiple destination nodes as the leaf nodes via intermediary nodes is known as a multicast tree or Steiner tree. The minimum Steiner tree, which is the tree that delivers optimal performance among all the trees, is the primary focus of this paper \cite{R13}.

Consider a graph $G=\left(V,E,c\right)$, where $V$ is the set of nodes in graph $G$, $E$ is the set of edges, and $c$ is the cost function of each edge. $n$ represents the number of nodes in graph $G$, $n=\left|V\right|$. $m$ represents the number of edges in graph $G$, $m=\left|E\right|$. $e_{ij}$ represents the edge from node $i\ $to node $j$, $e_{ij}\in E$. $S$ denotes the multicast node set, $S\subseteq V$. And $S=s\cup D$, where $s$ is the source node , $D$ is the set of destination nodes, and $D=\left\{d_1,d_2,\ldots\right\}$. $G^\prime$ represents the subgraph that is a subset of $G$.

The minimum Steiner tree is the minimum spanning tree $T$ of subgraph $G^\prime$ that includes the path between $s$ and $D$, where $s\in V^\prime,D\in\ V^\prime$, and the total cost of the tree, $\sum_{e\in T} c\left(e\right)$, is minimum. Here, $T=\left(V^\prime,E^\prime\right)$ and $G^\prime\subseteq G,\ \ S\subseteq\ V^\prime\subseteq V$, and $E^\prime\subseteq E$.

\subsection{Multicast problem decomposition}\label{sec3.2}
Multicast communication involves transmitting data from a single source node to multiple destination nodes via a tree structure with a unique path from the source node to each destination node. The tree structure has a unique path from the source node to each destination node, and each fork node has a unique path to a destination node that can be reached by passing through the fork node or another fork node before taking another unique path to the destination node. As a result, a path connecting all destination nodes can be established. We employ hierarchical reinforcement learning to abstract the process of constructing a multicast tree into two subproblems. The first subproblem is selecting the fork node and designating it as the new starting point. The second subproblem requires establishing the optimal one-to-one unicast path from the fork node to each destination node.

Hierarchical reinforcement learning (HRL) extends the Markov Decision Process (MDP) in Reinforcement Learning to a Semi-Markov Decision Process. According to the hierarchical deep Q-network (h-DQN) algorithm, the value function in RL is also generalized to consider goals and states, and the learning process is abstracted into sub-goal learning and control policy learning. The agent employs a two-layer structure consisting of a meta controller and an intrinsic controller \cite{R42}. 

\begin{figure}[h]
	\centering
	\includegraphics[width=0.9\linewidth]{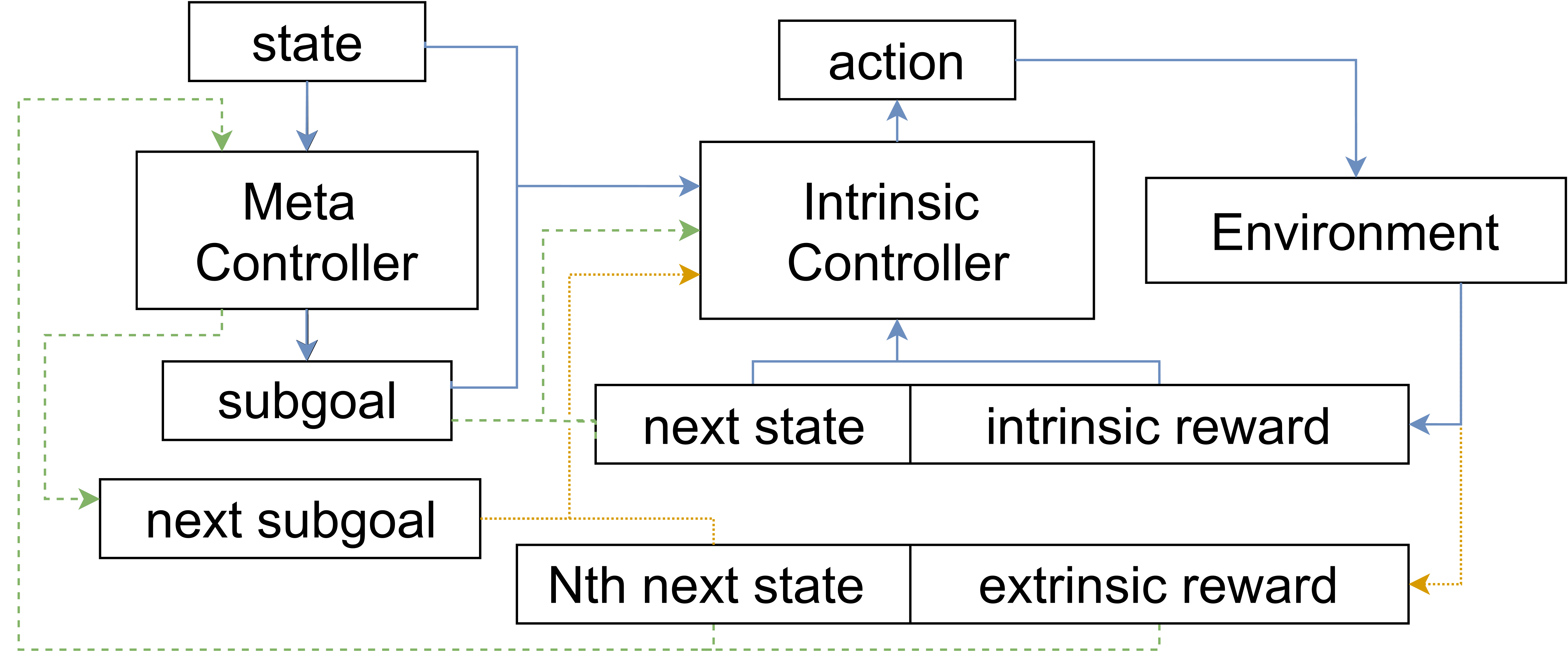}
	\caption{The structure of the h-DQN algorithm}
	\label{fig1}
\end{figure}

The meta-controller receives state $s_t$ as an input and outputs a sub-goal $g_t\in\mathcal{G}$, where $\mathcal{G}$ is a set of all available sub-goals at time $t$. The intrinsic-controller takes state $s_t$ and sub-goal $g_t$ as inputs to produce an action $a_t$ as an output. Through the blue solid line shown in Fig.\ref{fig1}, the agent interacts with the environment by carrying out the intended action $a_t$, which provides the agent with the internal reward value $r_{t+1}^{in}$, and leads to being in the next state $s_{t+1}$. Afterward, the intrinsic-controller utilizes the green dashed line shown in Fig.\ref{fig1} to choose the next action $a_{t+1}$ based on the next state $s_{t+1}$ and the sub-goal $g_t$. The iterative process continues until the N-th step when the sub-goal is accomplished, by gathering the external reward value $r_{t+N+1}^{ex}$, which ultimately leads to entering the state $s_{t+N+1}$. The meta-controller chooses the next sub-goal $g_{t+N+1}$ based on state $s_{t+N+1}$ to make the iteration continue until the agent achieves the maximum reward value, represented by the orange dashed line shown in Fig.\ref{fig1}.

The meta-controller is responsible for selecting a fork node $N_f$ from a set of nodes $V$. In contrast, the intrinsic controller is responsible for selecting the next hop node as the action starting from $N_f$. The agent continues to make these selections until it discovers a unicast path represented by $P_{fd}$, connecting $N_f$ to a destination node $N_d$, where $N_d$ and $N_f$ are elements of $V$ and destination set $D$, respectively. By repeatedly doing this, the agent ultimately constructs a multicast tree, $T$, represented by a set of unicast paths, $P_{f_1d_1},P_{f_2d_2},\ldots,P_{f_kd_k}$. Here, $f_k$ represents the kth fork node selected, while $d_k$ represents the kth destination node.

\begin{figure}[h]
	\centering
	\includegraphics[width=0.7\linewidth]{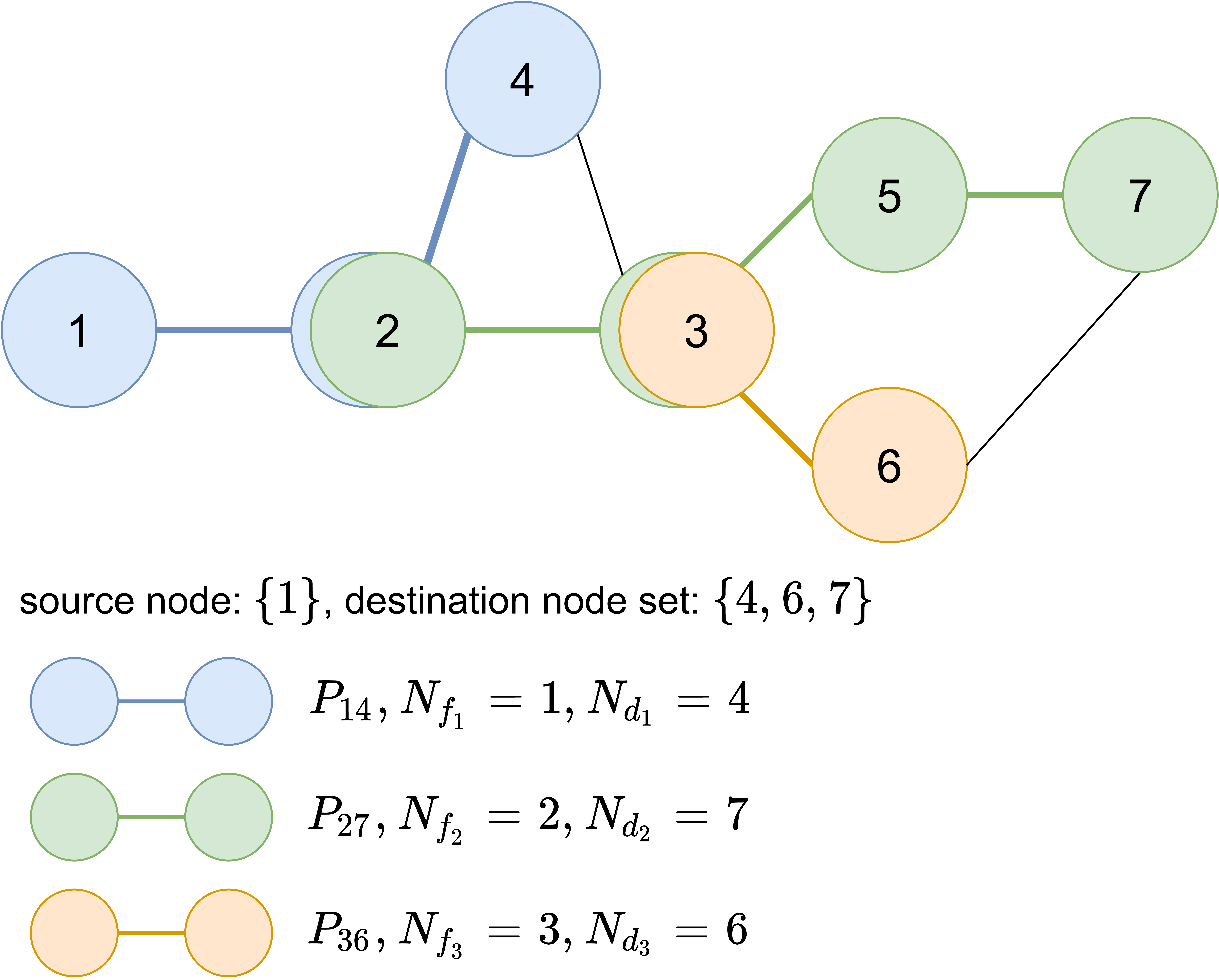}
	\caption{Constructing a multicast tree using fork nodes}
	\label{fig2}
\end{figure}

Fig.\ref{fig2} illustrates the construction of a multicast tree $T$ with the source node $s=1$and the destination node set $D=\{4,6,7\}$. The first branching node on the path from $N_1$ to $N_4$ is the initial node $N_1$ ($N_{f_1}=1$), and the destination node is $N_4$, making this path $P_{14}$. The second branching node is $N_2$ ($N_{f_2}=2$), and the destination node is $N_7$, yielding $P_{27}$. Finally, the third branching node is $N_3$ ($N_{f_3}=3$), and the destination node is $N_6$, resulting in $P_{36}$. Hence, we construct the multicast tree, $T=\{P_{14},P_{27},P_{36}\}$.

The optimization objective of the meta-controller is to maximize the average bandwidth ($bw_{tree}$), minimize the cumulative delay ($delay_{tree}$), and packet loss rate ($loss_{tree}$) of the tree. As these objectives may not be mutually independent, and they are be modeled as a multi-objective optimization problem, as shown in Eq.\ref{eq1}:

\begin{equation}\label{eq1}
\max F(T) = ({f_1}(T),{f_2}(T),{f_3}(T)){\rm{ }}
\end{equation}
Where $f_1\left(\mathrm{T}\right)$, $f_2\left(T\right)$ and $f_3\left(T\right)$ are shown in Eq.\ref{eq2}, Eq.\ref{eq3}, Eq.\ref{eq4} respectively:

\begin{equation}\label{eq2}
{f_1}(T) = b{w_{tree}} = average(\sum\limits_{d \in D} {\mathop {\min }\limits_{{e_{ij}} \in {p_{sd}}} (b{w_{ij}})} )
\end{equation}

\begin{equation}\label{eq3}
{f_2}(T) = dela{y_{tree}} = \sum\limits_{{e_{ij}} \in T} {dela{y_{ij}}}
\end{equation}

\begin{equation}\label{eq4}
{f_3}(T) = los{s_{tree}} = 1 - \prod\limits_{{e_{ij}} \in T} {(1 - los{s_{ij}})}
\end{equation}
where $T$ is a Steiner tree in the graph $G$, $bw_{ij}$ represents the link bandwidth of $e_{ij}$, and $p_{sd}\in T$ represents the path in the multicast tree $T$ from the source node $s$ to the destination node $d$. $delay_{ij}$ denotes the link delay of $e_{ij}$. $loss_{ij}$ denotes the link packets loss rate of $e_{ij}$.

The optimization objective for the path $P_{fd}$ constructed by the intrinsic controller from the selected fork node $N_f$ by the meta controller to the destination node $N_d$ is as Eq.\ref{eq5}:
\begin{equation}\label{eq5}
\max F({{\rm{P}}_{fd}}) = ({f_1}({{\rm{P}}_{fd}}),{f_2}({{\rm{P}}_{fd}}),{f_3}({{\rm{P}}_{fd}})){\rm{ }}
\end{equation}
Where $f_1\left(\mathrm{P}_{fd}\right)$, $f_2\left(P_{fd}\right)$ and $f_3\left(P_{fd}\right)$ are shown in Eq.\ref{eq6}, Eq.\ref{eq7} and Eq.\ref{eq8} respectively:

\begin{equation}\label{eq6}
{f_1}({{\rm{P}}_{fd}}) = b{w_{{P_{fd}}}} = \mathop {\min }\limits_{{e_{ij}} \in {p_{fd}}} (b{w_{ij}})
\end{equation}

\begin{equation}\label{eq7}
{f_2}({P_{fd}}) = dela{y_{{P_{fd}}}} = \sum\limits_{{e_{ij}} \in {P_{fd}}} {dela{y_{ij}}}
\end{equation}

\begin{equation}\label{eq8}
{f_3}({P_{fd}}) = los{s_{{P_{fd}}}} = 1 - \prod\limits_{{e_{ij}} \in {P_{fd}}} {(1 - los{s_{ij}})}
\end{equation}
Thus, this paper introduces an algorithm, DHRL-FNMR, that uses deep hierarchical reinforcement learning. The algorithm finds paths through fork nodes for constructing multicast trees.

\section{Intelligent SDN multicast routing architecture}\label{sec4}
\begin{figure}[h]
	\centering
	\includegraphics[width=0.9\linewidth]{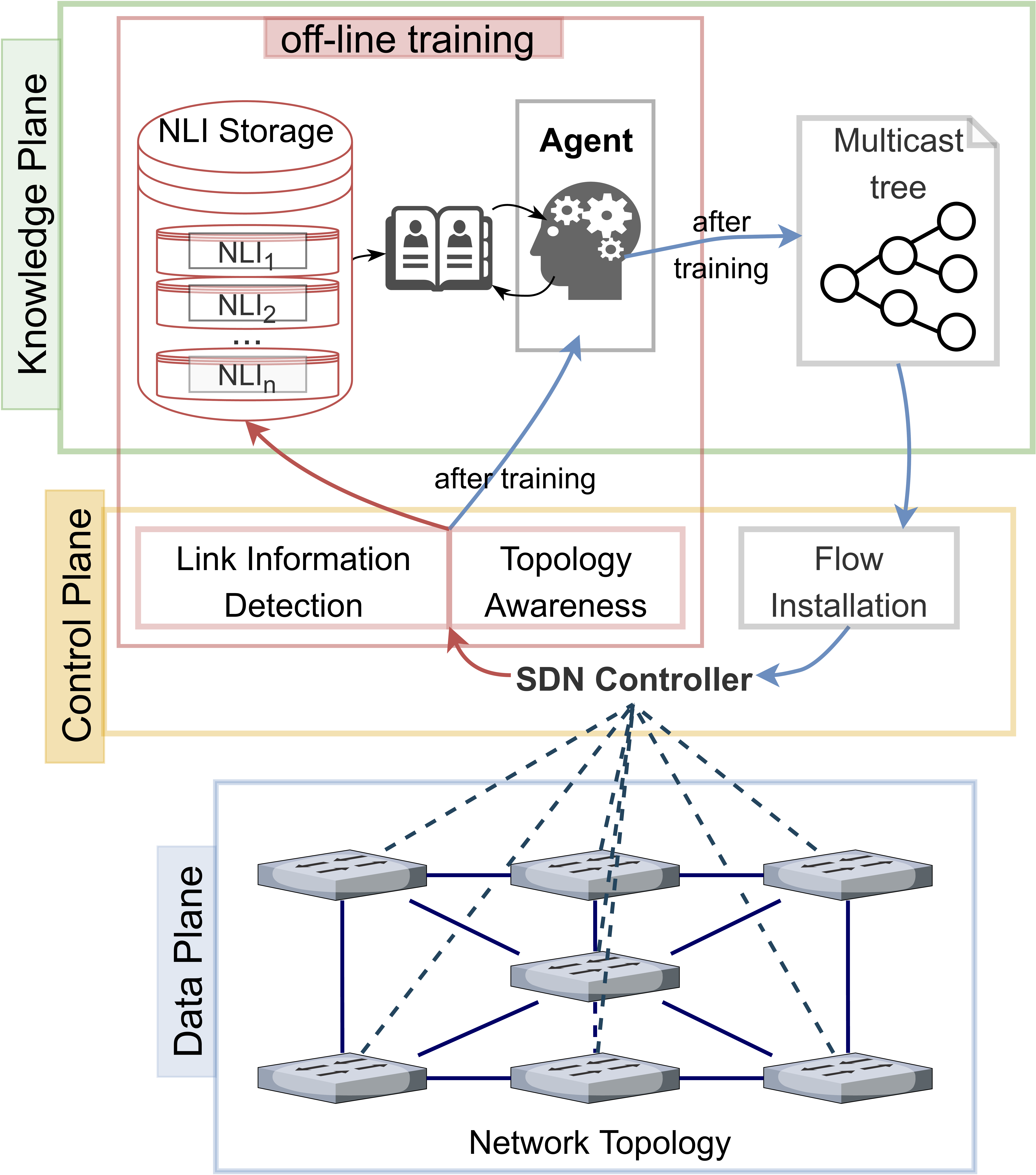}
	\caption{SDN intelligent multicast routing architecture}
	\label{fig3}
\end{figure}

This paper outlines an intelligent multicast routing framework for the DHRL-FNMR algorithm, and is depicted in Fig.\ref{fig3}. The framework is composed of three planes: the data plane, control plane, and knowledge plane, similar to the DRL-M4MR. In the control plane, an SDN controller interacts with the data plane using the OpenFlow protocol to gain network topology awareness, measure link information, and distribute flow tables. The gathered NLI (network link status information) is stored in the knowledge plane's NLI storage to support the DHRL-FNMR agent's training phase. Upon finishing the agent training, the SDN controller utilizes the current NLI to generate a multicast tree and distribute multicast routing flow tables accordingly. Further design details are explained below.

\subsection{Data plane}\label{sec4.1}
The data plane's OpenFlow switches use the OpenFlow protocol to connect and transmit packets to the SDN controller for routing and forwarding. Within each switch, incoming packets are evaluated against the internal flow table, and then forwarded to the output port. If the switch matches a multicast address, the packet will be forwarded to one or more ports based on the internal multicast routing flow table; otherwise, the packet will be discarded. Additionally, through the southbound interface, the data plane can respond to requests from the control plane.

\subsection{Control plane}\label{sec4.2}
The control plane encompasses a centralized SDN controller that connects to each switch in the data plane over the southbound interface, facilitating the gathering of global switch and link state insights, such as the remaining bandwidth, delay, and packet loss rate of each link. The SDN controller integrates three crucial functions: network topology awareness, link information detection, and flow table distribution.

Under network topology awareness, the SDN controller monitors switch state change events in the data plane, upon which it generates the network topology. Additionally, the SDN controller parses Address Resolution Protocol (ARP) packets to obtain host information, including Internet Protocol (IP) Addresses and Media Access Control (MAC) Addresses, and record the host's location.

The SDN controller measures the remaining bandwidth $bw_{ij}$, delay $delay_{ij}$, and packet loss rate $loss_{ij}$ of the links on the established network topology. To measure remaining bandwidth and packet loss rate of links, the SDN controller initiates a port information query request to the data plane, and subsequent to parsing the reply message, obtains statistics on number of bytes sent ($t_b$), bytes received ($r_b$), packets sent ($t_p$), packets received ($r_p$), and effective time ($t_{dur}$) for each port. The remaining bandwidth $bw_{ij}$ is deduced via Eq.\ref{eq9}, where $bw_c$ refers to the bandwidth link capacity. Eq.\ref{eq10} determines the packet loss rate $loss_{ij}$ of the link.

\begin{equation}\label{eq9}
b{w_{ij}} = b{w_c} - \frac{{\left| {\left( {{t_{b2}} + {r_{b2}}} \right) - \left( {{t_{b1}} + {r_{b1}}} \right)} \right|}}{{{t_{dur2}} - {t_{dur1}}}}
\end{equation}

\begin{equation}\label{eq10}
los{s_{ij}} = max\left( {\frac{{t{p_i} - r{p_j}}}{{t{p_i}}},\frac{{t{p_j} - r{p_i}}}{{t{p_j}}}} \right)
\end{equation}
where $t_{b\ast}$, $r_{b\ast}$, $t_{dur\ast}$ represent the number of bytes sent, the number of bytes received and the effective time at time $\ast$, respectively. $tp_\ast$, $rp_\ast$ denote the number of sent and received packets of port $\ast$, respectively.

Liao et al. proposed Eq.\ref{eq11} as an approximation for the $delay_{ij}$ of the link $e_{ij}$, which results from the measurement of delays of the two links that link each switch to the SDN controller, measured using ping requests and replies messages ($d_{echo1}$ and $d_{echo2}$) and the link delay between the switches, derived from LLDP packets ($d_{lldp1}$ and $d_{lldp2}$) \cite{R44}.

\begin{equation}\label{eq11}
dela{y_{ij}} = \frac{{\left( {{d_{lldp1}} + {d_{lldp2}} - {d_{echo1}} - {d_{echo2}}} \right)}}{2}
\end{equation}

The network link information (NLI) for the entire network topology is collected once every interval t and stored for future use.

One of the primary functions of the control plane involves deploying flow tables. Once the SDN controller has calculated either a unicast or multicast path, it can accordingly deploy the corresponding flow entries to the data plane routers, thus enabling data packets to be forwarded from input to output ports. If the destination node's address is a general IP address, the SDN controller deploys unicast flow tables. Conversely, if the address is a multicast address, multicast path flow entries are deployed. Additionally, switches located at the fork point of the multicast tree have flow entries set to forward data packets to different designated ports during the deployment of flow tables.

\subsection{knowledge plane}\label{sec4.3}
This paper's principal focus is on the knowledge plane, where the intelligent agent conducts offline training using collected network link state information to construct optimal multicast paths, which are then transmitted to the control plane. Additionally, the knowledge plane processes the network link status information (NLI) stored in the Network Link Information (NLI) storage, converting it into matrix input forms for the meta-controller and intrinsic-controller. State information for intrinsic and extrinsic controllers can be modified and diverse reward values can be provided as feedback in the programming environment. The model implementation and experience pool are also stored in the knowledge plane. A detailed algorithm design is presented in Section \ref{sec5}.

\section{Design of DHRL-FNMR}\label{sec5}
\begin{figure}[h]
	\centering
	\includegraphics[width=0.9\linewidth]{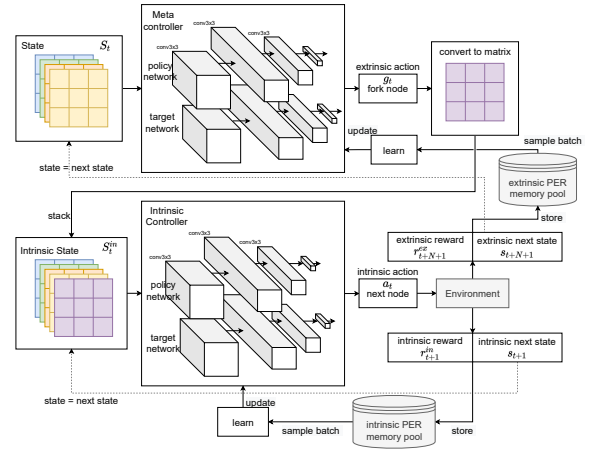}
	\caption{DHRL-FNMR algorithm framework}
	\label{fig4}
\end{figure}

The algorithm framework is designed in this article as shown in Fig.\ref{fig4}. Firstly, the outer meta-controller, in the current state $s_t$, i.e. the current multicast tree state matrix, remaining bandwidth matrix, link delay matrix, and packet loss rate matrix, selects a fork node as the sub-goal $g_t$ from all nodes based on the Q-value function. The Q-value function is fitted using a convolutional neural network, and a double network, i.e. a policy network and a target network, is used to reduce the impact of overestimation. 

Then, the fork node is transformed into a sub-goal matrix, stacked with the state $s_t$, and used as the state $s_t^{in}$ for the intrinsic controller. Based on $s_t^{in}$ and $g_t$, the intrinsic controller selects the next hop node as the action $a_t$. The agent interacts with the environment, and the reward function calculates the reward value $r_{t+1}^{in}$ and the next state $s_{t+1}$ for the intrinsic controller based on different actions. 

After storing the transition $\left(s_t,g_t,r_{t+1}^{in},s_{t+1}\right)$ in the internal prioritized experience replay pool, the agent collects a mini-batch from it for loss calculation, uses gradient descent to update the parameters $\theta_1$ of the internal policy network, and then proceeds to the next state iteration internally. When the agent reaches any of the destination nodes at step $N$, it is considered as completing a sub-goal, and the environment provides feedback on the next state $s_{t+N+1}$, and calculates the external reward value $r_{t+N+1}^{ex}$ through the reward function. 

The meta-controller's policy network is updated through batch sampling from the external prioritized experience replay pool. This process is similar to the intrinsic controller learning process. The external prioritized experience replay pool stores the transition tuple $\left(s_t,g_t,a_t,r_{t+N+1}^{ex},s_{t+N+1}\right)$. 

\begin{figure}[h]
	\centering
	\includegraphics[width=0.9\linewidth]{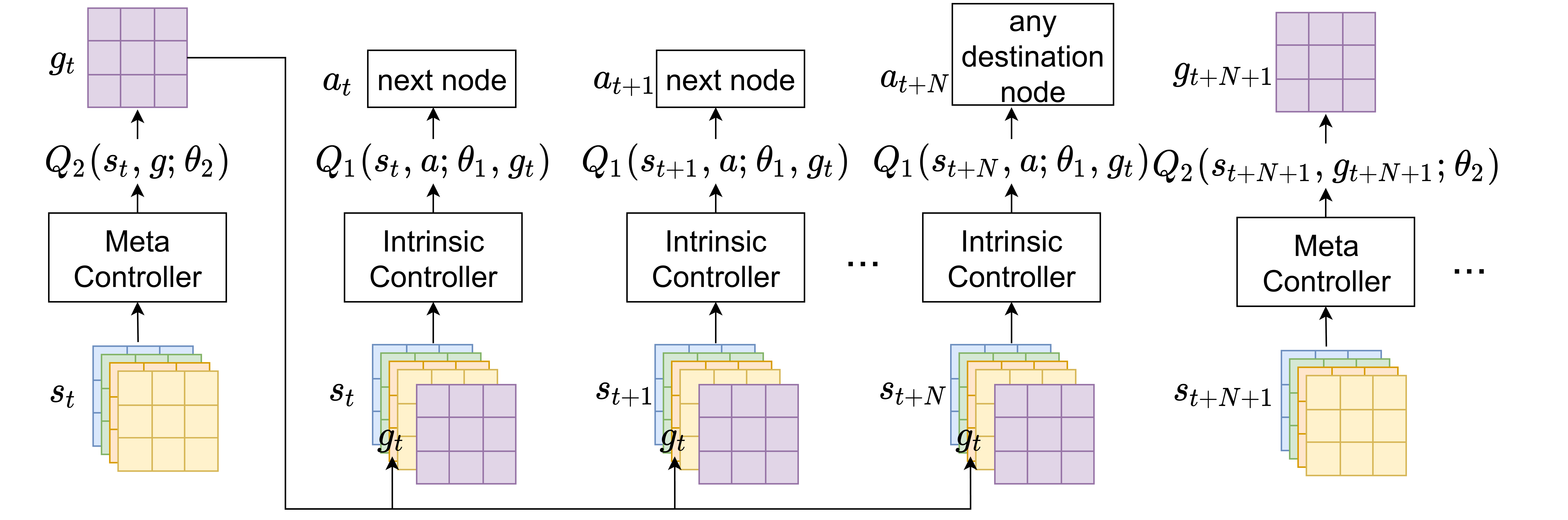}
	\caption{The Learning process of intrinsic and meta controllers}
	\label{fig5}
\end{figure}

After updating the meta-controller's policy network, the meta-controller selects a fork node, $g_{t+1}$, and the learning process continues iteratively until all destination nodes are reached, completing the agent's learning process. The learning process of both the intrinsic and meta controllers is shown in Fig.\ref{fig5}.

Finally, at fixed intervals, the agent updates the policy network parameters to the target network. Through training, the agent converges to the maximum reward value, which enables the construction of the optimal multicast tree.

\subsection{State space}\label{sec5.1}
\begin{figure}[h]
	\centering
	\includegraphics[width=0.8\linewidth]{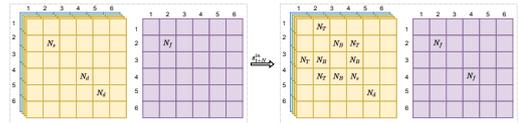}
	\caption{Design of state space, where matrices of different colors represent ${M}_{bw}$, ${M}_{delay}$, ${M}_{{loss}}$, and ${M}_{g}$}
	\label{fig6}
\end{figure}

The state space of the meta-controller consists of the link remaining bandwidth matrix $M_{bw}$, link delay matrix $M_{delay}$, and link loss rate matrix $M_{loss}$, which are designed in the same way as in DRL-M4MR, using normalized link information as elements of the matrix. In this paper, an additional channel is added as the sub-target matrix, and the design of the multicast tree state matrix $M_T$ is modified. For the $M_T$ matrix, whenever the agent's state moves to the next hop node, the diagonal element of the corresponding node in $M_T$ is marked as $N_s$ and the mark of the previous step is cancelled. When a new fork node is selected, it is marked as $N_s$, as shown in the yellow matrix in Fig.\ref{fig6}.

The state space of the intrinsic controller is expanded with a sub-target matrix $M_g$. Whenever the meta-controller selects a node as a fork node, the corresponding diagonal element in $M_g$ is marked as $N_f$. As shown in the purple matrix in Fig.\ref{fig6}, $N_2$ is first selected as the fork node, and then $N_3$ is selected. Whenever the same link or node is selected repeatedly, the markings are stacked to represent different states of the agent. The intrinsic controller uses a dimension stacking operation to stack the target matrix onto the channel dimension of the state space of the meta-controller, thus forming the input of the state-input neural network of the intrinsic controller.

\subsection{Action space}\label{sec5.2}
The two controllers in hierarchical reinforcement learning have different learning tasks and are designed with different action spaces, consisting of the external action space and the internal action space. As the output of the neural network is a tensor with a fixed dimension, each element of the tensor corresponds to a physical meaning. Translation: The external action space is the set of all nodes in the network, denoted by $\mathcal{A}_{ex}=V=\{N_1,N_2,\ldots,N_n\}$, where $n\ $is the number of nodes in the graph $G$. Each element of the tensor corresponds to a node in the graph. When making a decision, the meta-controller selects one of the $n$ nodes as its action, which also serves as the sub-goal for the intrinsic controller. This means that the agent selects this node as the fork node and passes the goal of finding the destination node starting from this node to the lower-level intrinsic controller. Since the output contains all the nodes in the graph, there are two possible scenarios when selecting a fork node:

(1) If the selected fork node is not in the current multicast tree, the state remains unchanged, and the agent is penalized before selecting a new action.

(2) If the selected fork node is a node in the current multicast tree, the intrinsic controller can continue to run.

Once any of the destination nodes is reached, a new node is selected as the fork node until all the destination nodes are found. When all the destination nodes are found, the extrinsic controller terminates the task, which also means the completion of the multicast tree construction. 

\begin{figure}[h]
	\centering
	\includegraphics[width=0.9\linewidth]{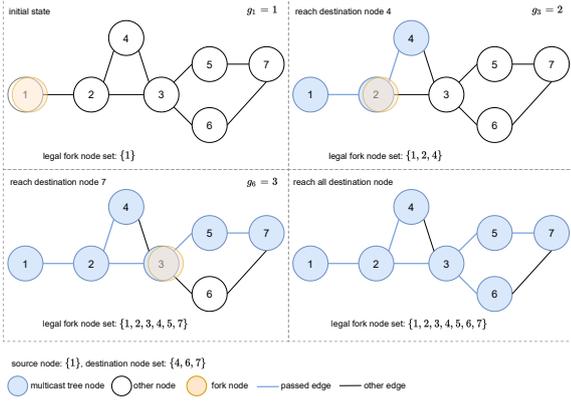}
	\caption{Design of meta controller action space}
	\label{fig7}
\end{figure}

Fig.\ref{fig7} illustrates an example of the actions selected by the meta-controller. The source node is $N_1$ and the destination node set is $N_4$, $N_6$, and $N_7$. The blue circles represent the multicast tree nodes that have been constructed, the orange circles represent the nodes that have been selected as fork nodes, the white circles represent other nodes in the graph, the blue lines represent the links of the multicast tree that have been constructed, and the black lines represent other links in the graph. When in the initial state, only the source node can be selected as the fork node. In this case, $N_1$ is selected as the fork node, i.e., $g_1=1$, and then the destination nodes are searched for. Suppose the intrinsic controller reaches the destination node $N_4$ after making two decisions. The meta-controller chooses a new fork node, and if $g\in\{1,2,4\}$, it is considered valid and the intrinsic controller continues to learn. Otherwise, it is punished. Suppose $N_2$ is chosen as the fork node, then $g_3=2$. The intrinsic controller continues to make decisions, and when it reaches the destination node $N_7$ after three steps, the meta-controller can choose from the sub-goals set $\{1,2,3,4,5,7\}$, and if $g\in\{1,2,3,4,5,7\}$, it is considered valid. Suppose $N_3$ is chosen as the fork node, then $g_6=3$. The intrinsic controller starts from $N_3$ and finds the destination node $N_6$. The agent no longer chooses the next round of action selection until all destination nodes have been reached.

The internal action space is denoted as $\mathcal{A}_{in}=EN_i=\{e_{ij_1},e_{ij_2},\ldots,e_{ij_l}\}$, where $EN_i$ represents the set of edges adjacent to the current node $N_i\in V$, and l denotes the maximum degree of graph $G$. Each element of the tensor corresponds to the adjacent edges of the current node, and is considered as None when there is no corresponding edge. After each action selection by the agent and interaction with the environment, there will be four possible scenarios:

(1) If the agent selects an edge that can be used for the next hop, the current state will be changed to the next state $S_{t+1}$, and a positive reward will be obtained.

(2) If the agent selects an edge that has no corresponding edge, the current state will not be changed, and an invalid selection penalty will be obtained.

(3) If the agent selects an edge that has already been included in the multicast tree, the corresponding flag will be added and the agent will enter the next state, while receiving a loop penalty.

(4) If the selected edge forms a cycle, the agent will enter the next state and receive a cycle penalty.

\begin{figure}[h]
	\centering
	\includegraphics[width=0.9\linewidth]{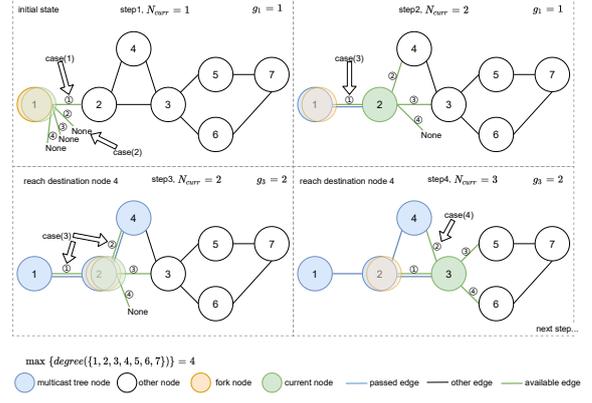}
	\caption{Design of intrinsic controller action space}
	\label{fig8}
\end{figure}
Fig.\ref{fig8} shows a graph with $N_1$ as the source node and $N_4$, $N_6$, and $N_7$ as the set of destination nodes.  The nodes and links are color-coded: blue circles and lines represent nodes and links already in the multicast tree, orange circles represent fork nodes, green circles and lines represent the agent's current location and available links, and white circles and black lines represent other nodes and links in the graph.  Since the maximum degree in the graph is 4, the neural network outputs a tensor of dimension $\left(1\times4\right)$, with each element corresponding to one of the four adjacent edges to the current node.

For the initial state, the agent is at node $N_1$, with only one available edge, $e_{12}$, and the other three elements in the tensor are None. If the agent selects the first element of the tensor, i.e., $e_{12}$, as the action, it corresponds to the situation (1) mentioned above, and the current node of the agent will change to $N_2$; if other edges are selected, it corresponds to situation (2). 

In the second step, the current node is $N_2$. The four elements output by the tensor represent edges $e_{21}$, $e_{24}$, $e_{23}$, and None. Since edge $e_{21}$ is already in the multicast tree, selecting $e_{21}$ as the action corresponds to situation (3) mentioned above. As the agent makes decisions, the current node changes with each decision. 

In the fourth step, the current node is $N_3$. At this time, the four dimensions of the output tensor represent edges $e_{32}$, $e_{32}$, $e_{35}$, and $e_{36}$, respectively, and the four elements correspond to the four edges. If the agent selects $e_{34}$ as the action, it is considered as situation (4) mentioned above.

\subsection{Reward function}\label{sec5.3}
The reward function determines the learning direction of the agent, and we hope that the external meta-controller can learn the optimal fork node selection strategy, while the intrinsic controller can learn the optimal path strategy from a fork node to a specific destination node.
The external reward function is divided into two types, completing sub-goals and selecting illegal fork nodes:

(1) If the agent reaches a certain destination node from the fork node, it will receive a reward value$R_g,$ as shown in Eq.\ref{eq12}. 
\begin{equation}\label{eq12}
{R_g} ={\beta_1}bw_{P_{ij}} + {\beta_2}\left(1 - delay_{P_{ij}} \right) + {\beta_3}\left(1 - loss_{P_{ij}}\right)
\end{equation}
where, $P_{ij}$ represents the unicast path from fork node $N_i$ to destination node $d_j$. $bw_{P_{ij}}$ represents the bottleneck bandwidth of the path, which is calculated in the same way as $bw_{tree}$. $delay_{P_{ij}}$ represents the delay of the path, which is calculated in the same way as $delay_{tree}$. $loss_{P_{ij}}$ represents the product of the path's packet loss rate, which is calculated in the same way as $loss_{tree}$.
	
If the agent reaches all destination nodes, it will receive $R_{finish}$ calculated using Eq.\ref{eq13}.
\begin{equation}\label{eq13}
	\begin{split}
	R_{finish} 
	& = {\beta_1}bw_{tree} + {\beta_2}(1 - delay_{tree})\\
	& + {\beta_3}(1 - loss_{tree})
	\end{split}
\end{equation}

(2) When the agent selects an illegal fork node, it will be given a fixed negative penalty value, $R_{illegal}=C_1$.

If the sub-goal is not completed, that is, neither of the above two situations are satisfied, then there is no reward value, and the intrinsic controller learns at this time. 

The internal reward value is generated based on the interaction between the agent and the environment. We set the internal reward function to four types based on the state of the multicast tree that may be formed after the action is executed. 

(1) If the agent selects the correct next-hop node and the next-hop node is not a destination node, then it will receive a reward value $R_{step}$, which is calculated in the same way as DRL-M4MR, as shown in Eq.\ref{eq14}.

\begin{equation}\label{eq14}
{R_{step}} = {\beta_1}bw_{ij} + {\beta_2}\left(1 - delay_{ij}\right) + {\beta_3}\left(1 - loss_{ij}\right)
\end{equation}

(2) If there is no corresponding link for this action, then a constant penalty value $R_{none}=C_2$ is given. 

(3) If a link is selected as the next-hop link, but it already exists in the multicast tree or selecting it would cause a loop, then a constant penalty value $R_{back}=C_3$ is given. 

(4) If the agent reaches any destination node after selecting the link, it is considered as completing the subtask, and the final reward value $R_g$ is calculated and returned based on the path $P_{fd}$ taken.

In the above reward functions, to prevent any factor from having too much impact on the reward value, we use the same standardization method as DRL-M4MR.

\subsection{Q value function}\label{sec5.4}
The hierarchical reinforcement learning algorithm used in this paper is h-DQN, which is an improvement based on DQN and max-Q. The intrinsic controller of h-DQN adds a parameter target $g$ compared to traditional DQN, and evaluates the Q value according to Eq.\ref{eq15}:
\begin{equation}\label{eq15}
	\begin{split}
	&Q_1^*(s,a;g) 
	=\mathop {\max }\limits_{_{{\pi _{ag}}}} E[\sum\limits_{k = 0}^\infty  {{\gamma ^k}r_{t + k + 1}^{in}|s, a, g,{\pi _{ag}}} ]\\
	&= \mathop {\max }\limits_{_{{\pi _{ag}}}} E[r_{t + 1}^{in}
	+ \gamma \mathop {\max }\limits_{_{{\pi _{ag}}}} Q_1^*({s_{t + 1}},{a_{t + 1}};g)|s, a, g,{\pi _{ag}}]
	\end{split}
\end{equation}
where $g$ represents the target of the agent in state $s$, and $\pi_{ag}$ represents the executed action policy. This equation evaluates the value of executing the action in the current state under the current policy. The $Q$ value function of the meta-controller is similar to DQN, as shown in Eq.\ref{eq16}:
\begin{equation}\label{eq16}
	\begin{split}
	Q_2^*(s,g) = &\mathop {\max }\limits_{{\pi _g}} E[\sum\limits_{k = 0}^N r_{t + k + 1}^{ex} \\
	&+ \gamma \mathop {\max }\limits_{g\prime } Q_2^*({s_{t + N}},g')|s, g,{\pi_g}]
	\end{split}
\end{equation}
where $N$ represents the number of time steps for the controller to complete the given target, $g^\prime$ represents the target of the agent in state $s_{t+N}$, and $\pi_g$represents the policy under the current target.

Therefore, the transition obtained by the meta controller based on $Q_2$ is $\left(s_t,g_t,r_{t+N}^{ex},s_{t+N}\right)$, which is stored in the experience pool $PER_2$. On the other hand, the transition obtained by the intrinsic controller based on $Q_1$ is $\left(s_t,a_t,g_t,r_{t+1}^{in},s_{t+1}\right)$, which is stored in $PER_1$.
The loss function used to update the parameters of $Q_1$ and $Q_2$ is the Huber loss function, as shown in Eq.\ref{eq17} and Eq.\ref{eq18}:
\begin{equation}\label{eq17}
{L_n} = \left\{ {\begin{array}{*{20}{c}}
	{0.5{{({x_n} - {y_n})}^2},{\rm{    }}if\left| {{x_n} - {y_n}} \right| < 1}\\
	{|{x_n} - {y_n}| - 0.5,{\rm{    }}otherwise}
	\end{array}} \right.
\end{equation}

\begin{equation}\label{eq18}
{y_n} = r + \gamma ma{x_{a\prime }}Q\left( {s\prime ,a\prime ;\theta ,g} \right)
\end{equation}
where, $s^\prime$ and $a^\prime$ denote the next state and next action, respectively. For $Q_1$, $x_n$ is the predicted $Q_1$ value and $y_n$ is the expected $Q_1$ value for the nth sample in the mini-batch, and the same applies for $Q_2$.

\subsection{Exploration method}\label{sec5.5}
The Decay $\varepsilon$-greedy method is also used as the exploration strategy. In the early stages of training, the agent tries to sample different actions in the action space to explore and learn more experiences, as shown in Eq.\ref{eq19}.
\begin{equation}\label{eq19}
{a_t} = \left\{ {\begin{array}{*{20}{c}}
	{argma{x_a}{Q_t}\left( {s,a} \right),{\rm{   }}if{\rm{  }}x \ge \varepsilon }\\
	{random\ choice, otherwise}
	\end{array}} \right.
\end{equation}
where $x$ follows a uniform distribution between 0 and 1, denoted as $x\sim U\left(0,1\right)$. The value of $\varepsilon$ is calculated according to Eq.\ref{eq20}:
\begin{equation}\label{eq20}
\varepsilon  = {\varepsilon _{final}} + \left( {{\varepsilon _{start}} - {\varepsilon _{final}}} \right) \cdot exp\left( { - \frac{{epoc{h_{curr}}}}{{epoc{h_{decay}}}}} \right)
\end{equation}
where $\varepsilon_{start}$ represents the initial value of $\varepsilon$ at the beginning of the training. $epoch_{curr}$ represents the current epoch or training iteration, while $epoch_{decay}$ controls the convergence epoch, where smaller values lead to earlier convergence. As training progresses, $\varepsilon$ approaches $\varepsilon_{final}$, and both $\varepsilon_{start}$ and $\varepsilon_{final}$ are in the range $\left[0,1\right]$.

\subsection{Prioritized experience replay}\label{sec5.6}
The experience replay buffer uses the same prioritized experience replay as DRL-M4MR. Compared to uniform sampling from a typical experience buffer, in a prioritized experience replay buffer, transitions with a larger temporal difference error (TD-error) have a higher probability of being sampled. The agent learns more from these transitions, as they are more likely to contain valuable learning experiences \cite{R45}. The prioritized experience replay allows the agent to more easily sample useful learning samples instead of obtaining a large number of negative samples with penalty values, making the learning process more effective.

The probability of sampling is proportional to the TD error, i.e., $i~P\left(i\right)\propto\left|TD_{error}\right|^\alpha$, as shown in Eq.\ref{eq21}, where $\alpha$ is a hyperparameter that determines the shape of the distribution, and when $\alpha=0$, it corresponds to uniform sampling. Where, $p_i$ is a priority index determined by the TD error. Finally, the importance sampling parameter, $\omega_i$, is used to correct the distribution bias, as shown in Eq.\ref{eq22}, where $N$ is the size of the experience pool, and $\beta$ is the importance sampling hyperparameter.
\begin{equation}\label{eq21}
P(i) = \frac{{{p_i}^\alpha }}{{\sum\nolimits_k {p_k^a} }}
\end{equation}

\begin{equation}\label{eq22}
{\omega _i} = {(N \cdot P(i))^{ - \beta }}
\end{equation}

\subsection{h-DQN based multicast routing algorithm}\label{sec5.7}

The agent of DHRL-FNMR is trained based on the collected network environment information, using the input of source node $N_s$ and destination node set $N_D$ to find the maximum reward value as the target, in order to construct the optimal multicast tree. As shown in Alg.\ref{alg1}, the input of the agent includes: source node $N_s$, destination node set $N_D$, total training epochs $M$, hyperparameters of intrinsic and extrinsic controllers such as learning rate $\alpha$, batch size $k$ for sampling, target network parameter update frequency $n_{update}$, convergence epochs of Decay $\varepsilon$-greedy, and hyperparameters of intrinsic and extrinsic PER. The output is the multicast tree $T$ in the graph $G$ from $N_s$ to $N_D$.

Lines 1 to 3 are for model parameter initialization and PER experience pool initialization. The parameters of the policy networks for the inner controller and meta-controller are initialized first, where $\theta_1$ and $\theta_2$ are the neural network parameters for the policy of the inner controller and meta-controller, respectively. Then, the neural network parameters of the target networks, ${\hat{\theta}}_1$ and ${\hat{\theta}}_2$, are initialized to the corresponding parameters of their policy networks. The inner and outer priority experience replay pools are then initialized based on the size of the experience pool, $M_m$.

After initialization, the agent begins training for a total of $M$ iterations, as shown in lines 4-43. Lines 5-7 show that the remaining bandwidth, delay, and loss rate of a network link are obtained from the network link status information repository for each time step, and are then converted into matrix form for the state space as $M_{BW}$, $M_{delay}$, and $M_{loss}$, respectively. The multicast tree status matrix $M_T$ for the initialized environment is then stacked in the channel dimension to form a matrix of size $\left(1\times4\times n\times n\right)$, which serves as the current state $s_t$ for the meta-controller.

Lines 8-41 depict the outer meta-controller loop. First, in line 9, the meta-controller obtains the sub-goal $g_t$ based on Decay $\varepsilon$-greedy at the current state $s_t$. In line 10, the environment determines if $g_t$ is a valid sub-goal and, if so, it is converted to matrix form $M_{g_t}$ and enters the intrinsic controller training loop from lines 11-33.

\begin{algorithm}[t]\footnotesize
	\renewcommand{\algorithmicrequire}{\textbf{Input:}}
	\renewcommand{\algorithmicensure}{\textbf{Output:}}
	\caption{HDRL-FNMR}\label{alg1}
	\begin{algorithmic}[1]
		\Require source node $N_s$, destination nodes set $N_D$, learning rate $\alpha$, reward decacy $\gamma$, batch size $k$, update frequency $n_{update}$, training episodes $M$;
		\Ensure optimal multicast tree for $(N_s, N_D)$;
		\State Initialize intrinsic controller's and meta controller's policy network with random weight $\theta_1$ and $\theta_2$;
		\State Initialize intrinsic controller's and meta controller's target network with weight $\hat{\theta_1}=\theta_1$, $\hat{\theta_2}=\theta_2$;
		\State Initialize PER pool $PER_1$ and $PER_2$;
		\For{$episode=1\leftarrow M$}
		\For{$M_{BW},M_{delay},M_{loss}$ in NLIs Storage}
		\State Reset environment with $(N_s, N_d)$ ;
		\State Get $s_t = stack(M_T,M_{BW},M_{delay},M_{loss})$;
		\While{True} \hspace*{5em} // meta-controller loop.
		\State $g_t = m\_c.sample(s_t)$; 
		\If{$env.subgoal\_step(g_t)$} \hspace*{2em} // $g_t$ is legal.
		\While{True} \hspace*{1em} // intrinsic controller loop.
		\State $s_t^{in} = stack(s_t, M_{g_t})$;
		\State $a_t = in\_c.sample(s_t^{in})$;
		\State $s_{t+1}, r_{t+1}^{in}, r_{t+1}^{ex}, s_f, g_f = env.step(a_t)$;
		\State $in\_c.store(s_t^{in}, a_t, r_{t+1}^{in}, s_{t+1})\ in\ PER_1$;
		\Statex \hspace*{7em} // $in\_c.learn()$ as follows.
		\State Sample batch $(s_i, a_i, r_{i+1}^{in}, s_{i+1})$ and get 
		\Statex \hspace*{7.5em} $\omega_i$ from $PER_1$, $i=1,2..,k$;
		\State Get $Q_1(s_i,a_i)$ from policy network;
		
		\If{$s_{i+1}$ is not None}
		\State $R' = r_{i+1}^{in} + \gamma max_{a'}\hat{Q_1}(s_{i+1},a')$;
		\Else
		\State $R' = r_{i+1}^{in}$;
		\EndIf
		
		\State Compute TD-error $\delta_i = R'-Q_1(s_i,a_i)$;
		\State Update $PER_1$ transition priority by $\delta_i$;
		\State Update $\theta_1$ by gradient descent;
		
		\If{$g_f$ is True}  \hspace*{4em} // reach any.
		\State $m\_c.store(s_t, g_t, r_{t+1}^{ex}, s_{t+1})\ in\ PER_2$;
		\State $m\_c.learn()$;
		\State break;
		\EndIf
		
		\State $s_t=s_{t+1}$;
		\State In every $n_{update}$ steps, $\hat\theta_1 = \theta_1$, $\hat\theta_2 = \theta_2$;
		\EndWhile
		
		\If{$s_f$ is True}  // reach all destinations.
		\State break;
		\EndIf
		
		\Else
		\State $m\_c.store(s_t, g_t, R_{illegal},s_{t+1})$ in $PER_2$;
		\State $m\_c.learn()$, calculate loss and update $\theta_2$;
		\EndIf		
		\EndWhile
		\EndFor
		\EndFor
		\State Use final policy networks with parameter $\theta_1$ and $\theta_2$ to construct multicast tree from $N_s$ to $N_D$, the agent execute the max Q-value action for each decision.  
	\end{algorithmic}
\end{algorithm}

Lines 12-14: First, the state of the meta-controller $s_t$ and the sub-goal matrix $M_{g_t}$ are stacked along the channel dimension as input to the inner controller. The action $a_t$ is obtained based on Decay $\varepsilon$-greedy. In line 14, the agent interacts with the environment by executing the action $a_t$. The environment provides feedback in the form of the next state $s_{t+1}$, the reward value $r_{t+1}^{in}$ generated by the inner controller's interaction with the environment, the reward value $r_{t+1}^{ex}$ generated by the meta-controller's interaction with the environment, the state flag $s_f$ indicating the state after the intrinsic controller executes the action, and the sub-goal flag $g_f$ indicating whether the sub-goal has been achieved. In line 15, the transition generated by the inner controller is stored in the inner priority experience replay pool $PER_2$.

The inner controller then learns from the experience in lines 16-28. First, a batch of size $k$ is sampled from $PER_2$ and stacked to form a $\left(k\times4\times n\times n\right)$ input to the policy network to obtain the $Q_1$ value. The expected $Q$ value is calculated, followed by the TD-error. The priority of the transition in the priority experience pool is updated based on the TD-error. Then, the loss value is computed using the Huber loss function, and the internal network parameters are updated using gradient descent. The pseudocode for the learn() function of the intrinsic and meta controllers is the same.

Lines 26-30: If the sub-goal is achieved, the transition is stored in $PER_1$, and the meta-controller learns from the experience. Finally, the intrinsic controller loop is exited, and the agent searches for a new sub-goal to train on. Otherwise, in line 31, the agent moves to the next state and continues training. In line 33, the policy network parameters are updated to the target network every $n_{update}$ time steps. Line 34 is reached when the agent reaches all the destination nodes, and the meta-controller loop is exited to start the next training session.

If the sub-goal $g_t$ is not valid, lines 38-39 are executed. The meta-controller receives the reward value $R_{illegal}$, and the next state is set to None. The transition is stored in $PER_1$, and the network parameters are updated by computing the loss value.

Finally, after the agent completes its training, it selects the optimal $Q$ value from the inner and meta policy networks corresponding to the current state to construct the optimal multicast tree from $N_s$ to $N_D$.

\section{Experiments and evaluation}\label{sec6}
The experimental control server used in this study was Ubuntu 20.04.3, equipped with a GeForce RTX 3090 graphics card. The SDN controller employed was Ryu, and the data plane was simulated using Mininet 2.3.0. Ryu provided a southbound interface for communication with Mininet's Open vSwitches via OpenFlow 1.3. The traffic generator used was iperf3. Network information was collected using Python scripts and stored in Pickle format for ease of access. The implementation of hierarchical reinforcement learning and deep neural networks was carried out using Python 3.8 and PyTorch 1.9.0.

\begin{figure}[h]
	\centering
	\includegraphics[width=0.8\linewidth]{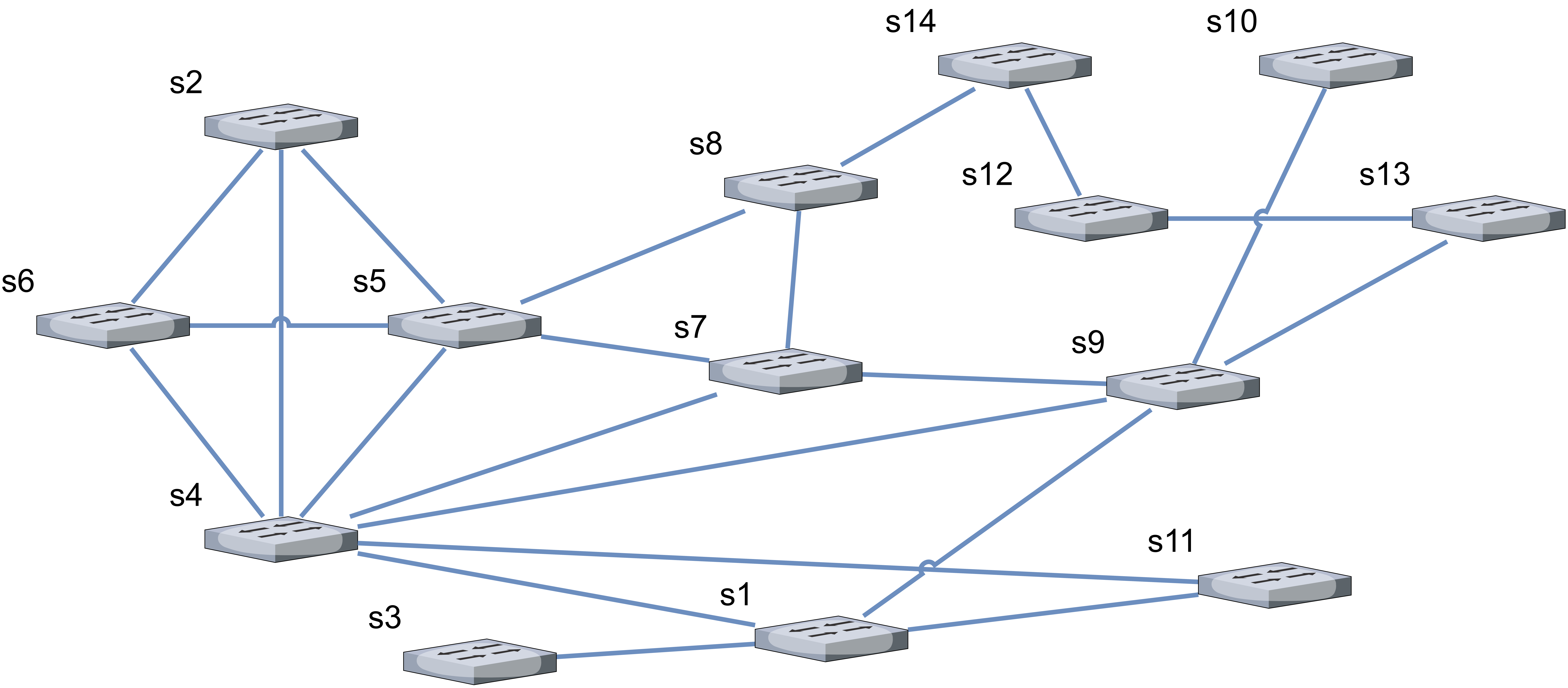}
	\caption{Experimental network topology}
	\label{fig9}
\end{figure}

\begin{figure}[h]
	\centering
	\includegraphics[width=0.8\linewidth]{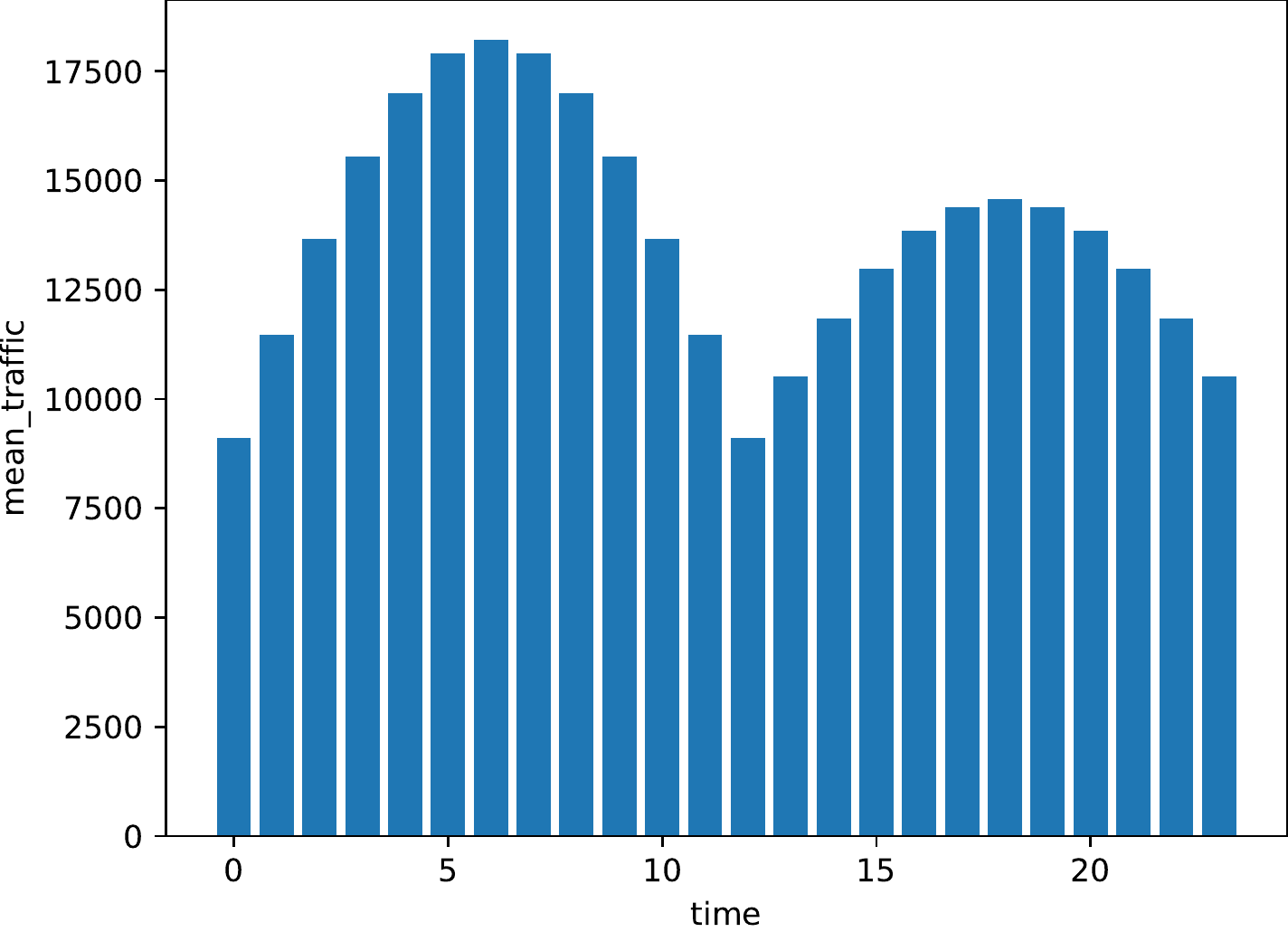}
	\caption{The average bandwidth of the traffic matrix}
	\label{fig10}
\end{figure}

The experimental topology is the same as that of DRL-M4MR, as shown in Fig.\ref{fig9}, and the link bandwidth and delay information are also the same. The traffic matrix is generated using the gravity model for traffic simulation, as shown in Fig.\ref{fig10}, and traffic is generated using the iperf3 tool by writing a Mininet script in Python.

\subsection{Parameter setting}\label{sec6.1}
The DHRL-FNMR agent consist of two layers of controllers, the neural network structure of the intrinsic and meta controllers are the same. They first pass through a three-layer convolutional neural network and then output through a fully connected layer. The input channel of the first convolutional neural network in the intrinsic controller is 5, the output channel is 128, the kernel size is $\left(3\times3\right)$, the stride is 1, and no padding is performed. The input channel of the second convolutional neural network is 128, the output channel is 256, the kernel size is $\left(3\times3\right)$, the stride is 1, and no padding is performed. The input channel of the third convolutional neural network is 256, the output channel is 128, the kernel size is $\left(3\times3\right)$, the stride is 1, and no padding is performed. The input channel of the fully connected layer is 8192, and the output channel is 7. The LeakyReLU activation function is applied after each convolutional layer. The input channel of the first convolutional neural network in the meta controller is 4, and the output channel of the final fully connected layer is 14. The rest of the network structure is the same as that of the intrinsic controller, as shown in the policy network and target network in Fig.\ref{fig4}. 

During the experiments, it was found that adjusting the kernel size had little impact on overall convergence, and the input and output channels had similarly minimal impact, as the input matrix contains a large number of 0 elements, and thus does not require a complex neural network to complete the feature extraction task.

First, we adjusted the parameters of the intrinsic controller. When adjusting the parameters of the intrinsic controller, we set the number of destination nodes to one, and the meta-controller always selects the source node as the fork node, making it easy to observe the convergence. In the experiment, we set S12 as the source node and S4 as the destination node. We evaluated the impact of parameter adjustments based on the reward value of the final constructed path $P_{12,4}$ by the agent. We chose S12 as the source node and S2, S4, and S11 as the destination nodes because we wanted the multicast tree from the source node to the destination node to have multiple paths, and we wanted the agent to face more link selection options when constructing the multicast tree to test the effectiveness of the algorithm. 

Through many experiments, we found that first, there is a leaf node around the destination node, and when the comprehensive parameters of the link on the leaf node, such as bandwidth, delay, and packet loss rate, are higher than those of the surrounding nodes, the agent will choose the redundant branch, so we chose S11. Secondly, there are multiple paths that can reach the destination node, and the number of steps to reach the destination node is not the same. For example, if we choose S2, we can reach it directly through $e_{\left(5,2\right)}$, or we can choose $e_{\left(5,4\right)}$ and then reach S2 through $e_{\left(4,2\right)}$, or we can reach S4 first and then reach S2 through S6. At this point, there is a situation where the comprehensive parameters of $e_{\left(5,4\right)}$ are higher than those of $e_{\left(5,2\right)}$, but the comprehensive parameters of $e_{\left(4,2\right)}$ are very low, resulting in the overall parameters of the tree being very low. We observed the agent's decision-making process in this case. There are seven links around S4, so there are multiple paths to reach the destination node, and we observed whether the agent could construct the optimal multicast tree. The hyperparameters of the DHRL part of the meta-controller were adjusted in the same way as those of the internal controller.

\begin{figure}[h]
	\centering
	\includegraphics[width=0.9\linewidth]{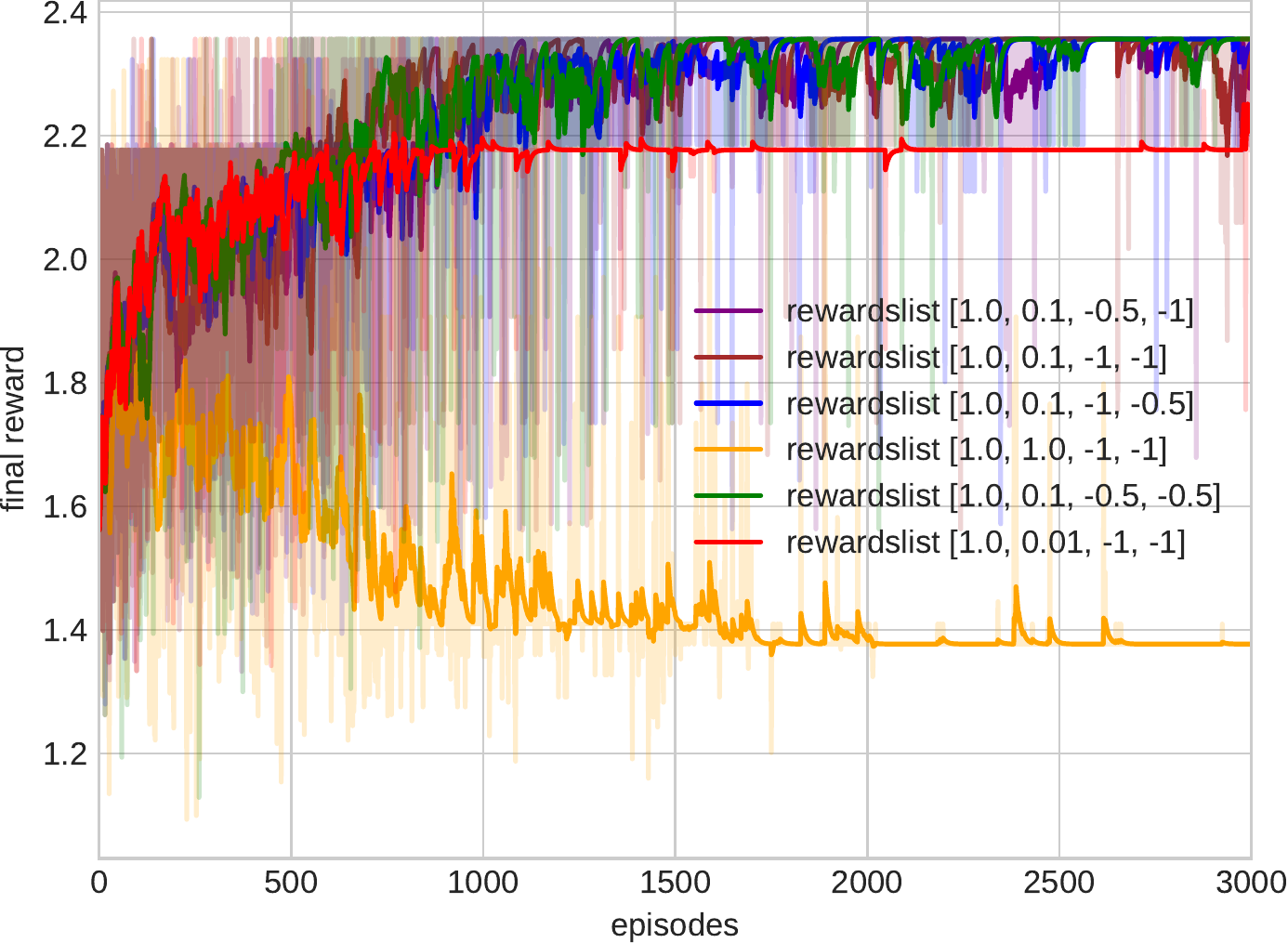}
	\caption{Reward value adjustment}
	\label{fig11}
\end{figure}

There have been numerous experimental adjustments made to the reward function setting. As shown in Fig.\ref{fig11}, the elements in the list represent the final reward function value $R_g$, the single-step reward function value $R_{step}$, the punishment for invalid selection $R_{none}$, and the punishment for looping $R_{back}$.

When $R_g$ is 1:1 compared to $R_{step}$, it can be seen from the figure that the path constructed by the agent did not converge to the optimal solution, and the path constructed during training was very unstable. This is because the reward value obtained for each step was very high, and the agent attempted to select more paths to obtain a higher reward value. During the experiment, it was found that the agent would select all the paths in the topology because the punishment value was low, even if the agent was punished, it would still choose more paths. Increasing the punishment value for looping would only reduce the situation where the agent walks back on the constructed path, and the agent would still construct a longer path.

When $R_g$ is 1:0.1 compared to $R_{step}$, the agent converges to a suboptimal solution. During the experiment, it was found that when the agent reached S5, there were two ways for the agent to reach S2: one was to go directly through $e_{\left(5,2\right)}$, and the other was to choose $e_{\left(5,4\right)}$ and then pass through $e_{\left(4,2\right)}$. At this time, the reward value for choosing $e_{\left(5,4\right)}$ was higher than that for $e_{\left(5,2\right)}$, but the reward value for choosing $e_{\left(4,2\right)}$ was very low after reaching S4. This is because the bandwidth of $e_{\left(4,2\right)}$ became the bottleneck bandwidth for the entire $P_{12,2}$, and the delay and packet loss rate were also relatively high, causing the $Q$ value for this link to increase after updating the parameters, and the agent became stuck in a suboptimal solution. In the first few steps of agent decision-making, when the single-step reward is high, it has a greater impact on decision-making. For example, when starting from S12, the reward value for $e_{\left(12,14\right)}$ is greater than that for $e_{\left(12,13\right)}$, which causes the agent to always choose $e_{\left(12,14\right)}$. However, the reward value for $e_{\left(14,8\right)}$ is very small, which affects the parameters for the entire path. In comparison, although the reward value for $e_{\left(12,13\right)}$ is slightly smaller, the subsequent reward values are higher than that for $e_{\left(14,8\right)}$. Therefore, combining the above two situations, the single-step reward value should be much smaller than the final reward value to ensure the convergence of the constructed path.

When $R_g$ is 1:0.01 compared to $R_{step}$ and single-step reward is not needed, the intrinsic controller of the agent can converge to the optimal solution. As shown in the figure, when $R_g$ is 1:0.01, the Q value of the link near the destination node is larger, while the Q value near the source node is very small when the agent constructs the path, and the final reward value becomes smaller and smaller as it updates backward, which leads to the problem of converging to a local optimum.

The penalty value has little effect on overall convergence, and when the penalty value is too large, it can easily lead to unstable convergence. As shown in Fig.\ref{fig11}, when both penalty values are -1, the agent produces strong fluctuations after convergence, while when both penalty values are -0.5, the fluctuations are smaller.

\begin{figure}[h]
	\centering
	\includegraphics[width=0.9\linewidth]{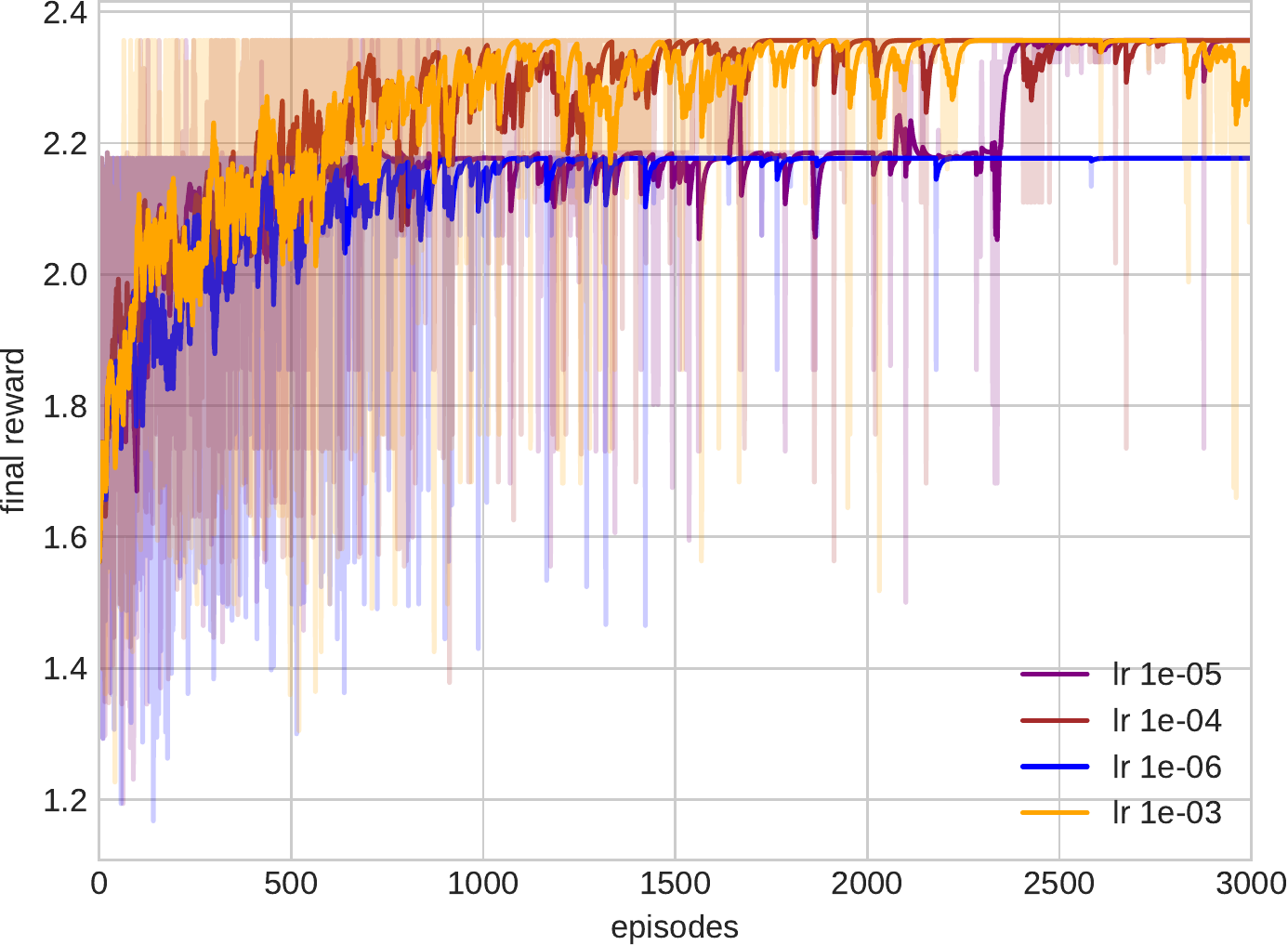}
	\caption{Learning rate adjustment}
	\label{fig12}
\end{figure}

The learning rate $\alpha$ is set to 1e-3, 1e-4, 1e-5, and 1e-6 respectively. As shown in Fig.\ref{fig12}, the vertical axis represents the final reward value of the intrinsic controller. Different learning rates of 1e-3, 1e-4, and 1e-5 all ensure that the agent converges to the optimal solution. When $\alpha$ is 1e-3, the agent converges the fastest, but is very unstable in the later stages of training. The curve fluctuates because in the double network, the Q value of the policy network will approach the Q value obtained by the target network, and the parameters of the target network will be iteratively updated based on the policy network, causing the loss to be already very low, but the parameters of the target network are updated and the loss value becomes very high, causing the parameters of the policy network to change greatly, leading to deviation in the agent's decision-making. When the learning rate is small at 1e-6, the curve converges to a local optimum.

\begin{figure}[h]
	\centering
	\includegraphics[width=0.9\linewidth]{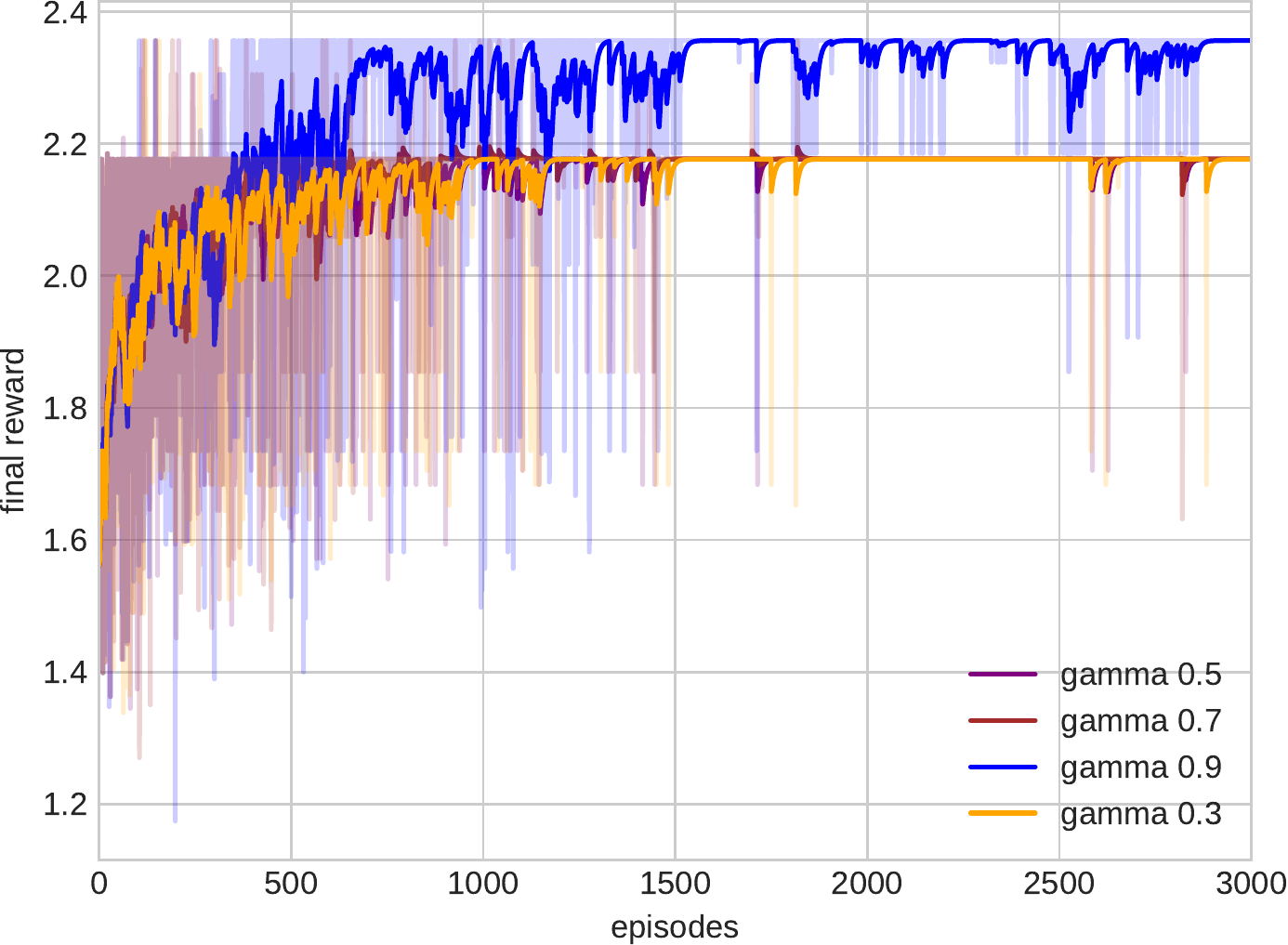}
	\caption{The discount factor adjustment}
	\label{fig13}
\end{figure}

The discount factor $\gamma$ is set to 0.3, 0.5, 0.7, and 0.9. As shown in Fig.\ref{fig13}, according to the final reward value, when the discount factor is small, the target reward value to be learned is reduced by the discount factor, and the agent cannot learn from the experience based on the current decision reward value. Therefore, according to the experimental results, a value of 0.9 is the best.

\begin{figure}[h]
	\centering
	\includegraphics[width=0.9\linewidth]{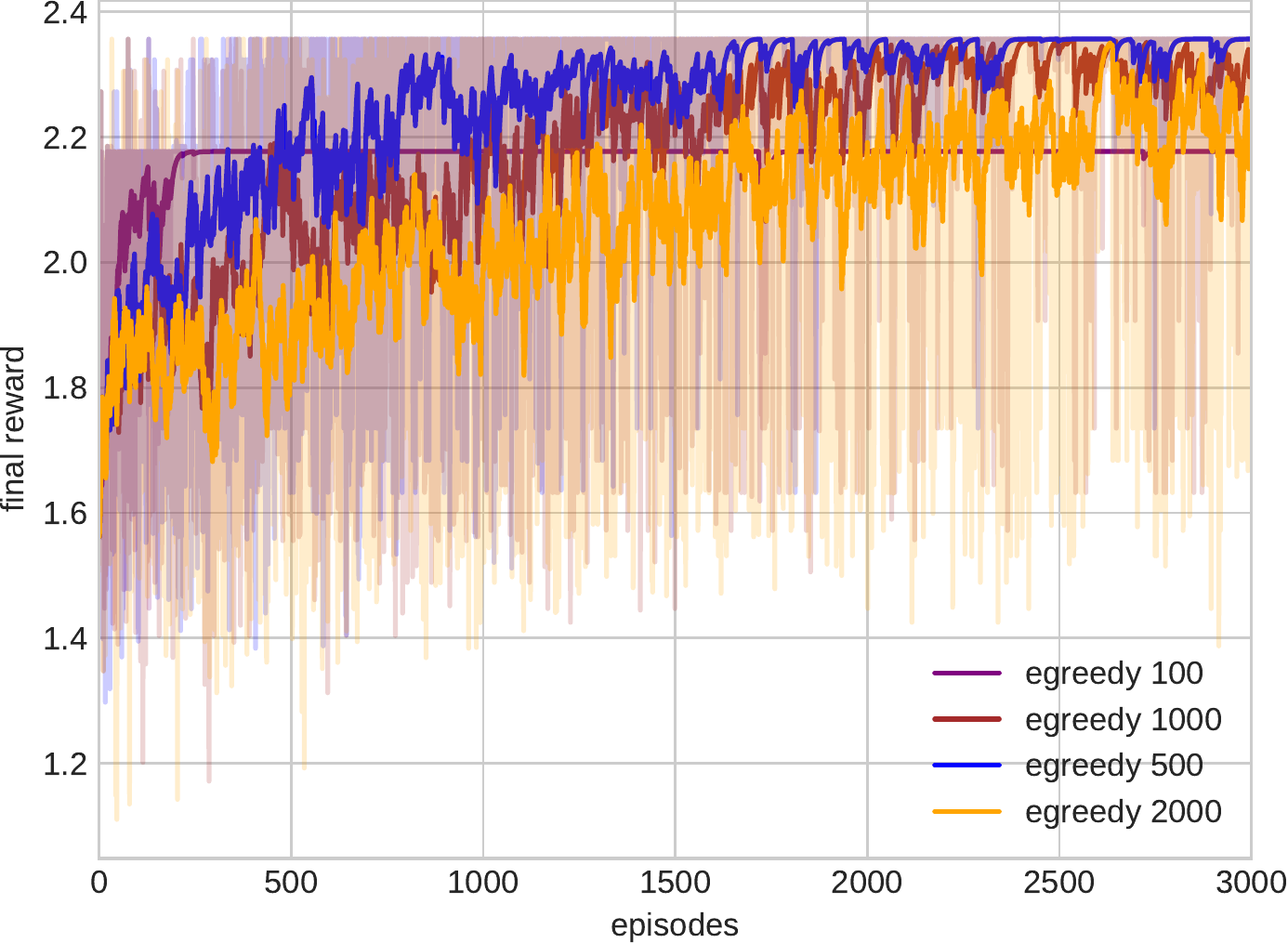}
	\caption{Adjustment of convergence epoch}
	\label{fig14}
\end{figure}

The convergence iteration of Decay $\varepsilon$-greedy is set to 100, 500, 1000, and 2000. As shown in Fig.\ref{fig14}, the larger the convergence iteration, the later the final reward value converges. When the convergence iteration is large, the agent will do a lot of exploration and randomly choose actions instead of selecting based on Q values, which leads to the agent continuously exploring instead of learning from the experience, as shown when it is set to 1000 and 2000. Although setting it to 100 converges the fastest, it also loses more exploration opportunities to find the optimal solution, leading to convergence to a local optimum. Therefore, to balance the relationship between exploration and exploitation, the convergence iteration is set to 500.

\begin{figure}[h]
	\centering
	\includegraphics[width=0.9\linewidth]{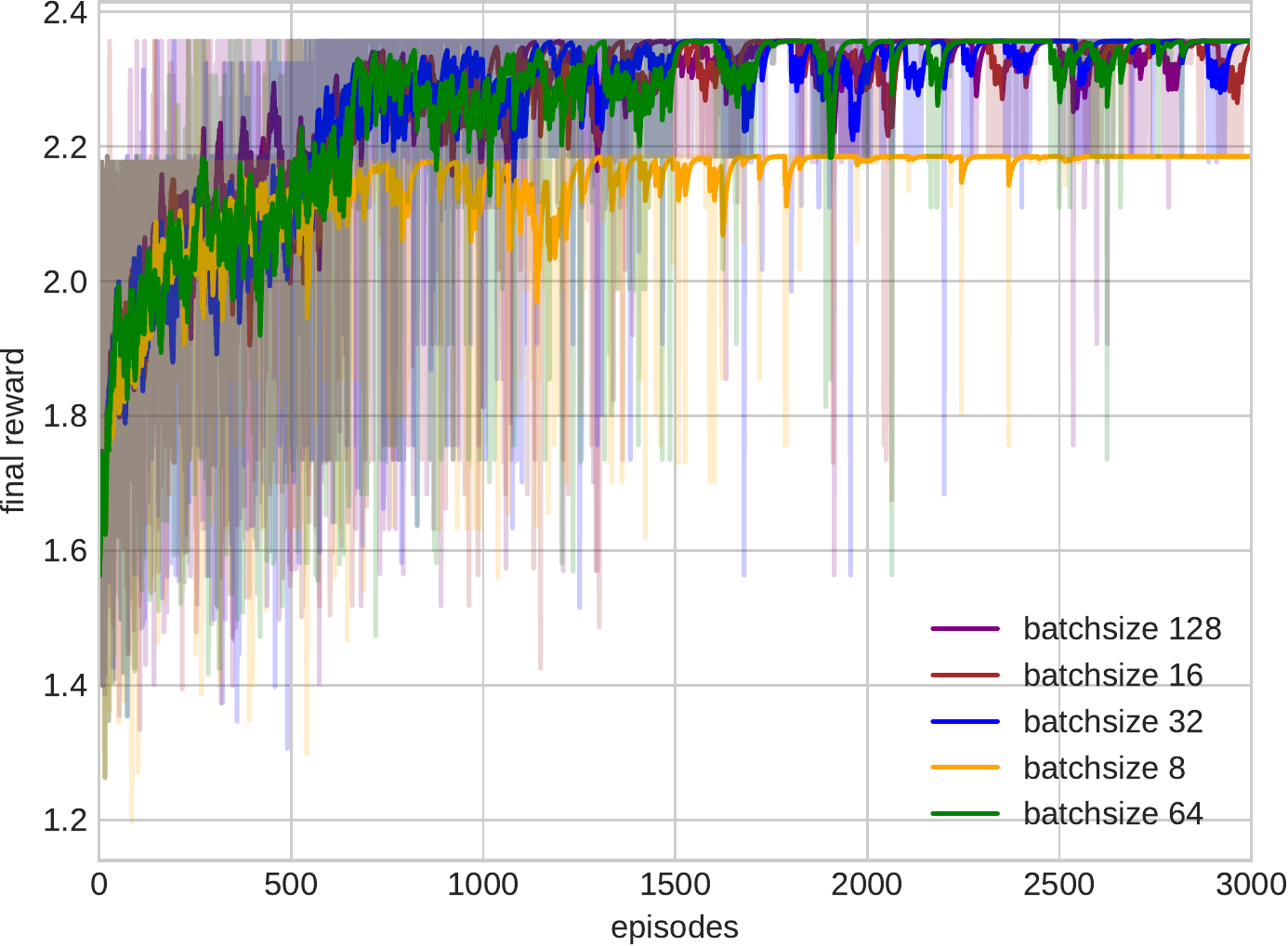}
	\caption{Adjustment of batch-size}
	\label{fig15}
\end{figure}

The batch-size refers to the number of transitions collected by an agent from the experience pool at once, which are set to 8, 16, 32, 64, and 128, with the experience pool size fixed at 2048, as shown in Fig.\ref{fig15}. Values of 6, 32, 64, and 128 can all make the agent converge to the optimal path, and there is not much difference between them. When the batch size is 8, the agent converges to a local optimum. This is because in the early stage of training, the agent encounters a large number of negative samples, such as loops, cycles, and invalid selections, resulting in fewer transitions that construct the path. Prioritized experience replay ensures the learning efficiency of the agent by importance sampling. After calculating the loss function for each sample in the batch size, they need to be averaged together, and then the loss is back-propagated. Therefore, when the experience pool contains a large number of positive samples and few negative samples in the later stage of learning, the negative samples have less impact after averaging, and the agent converges more stably with a larger batch size. However, the training time is also longer.

When the experience pool size is changed to 1e4, it is found that the agent is prone to converge to a local optimum when the experience pool is too large. Although the agent can discover better transitions, the probability of being collected is greatly reduced. In the early stage of exploration, only a few transitions with high reward values are discovered, and the experience pool is updated with many negative samples, which cannot effectively utilize the experience.

\begin{figure}[h]
	\centering
	\includegraphics[width=0.9\linewidth]{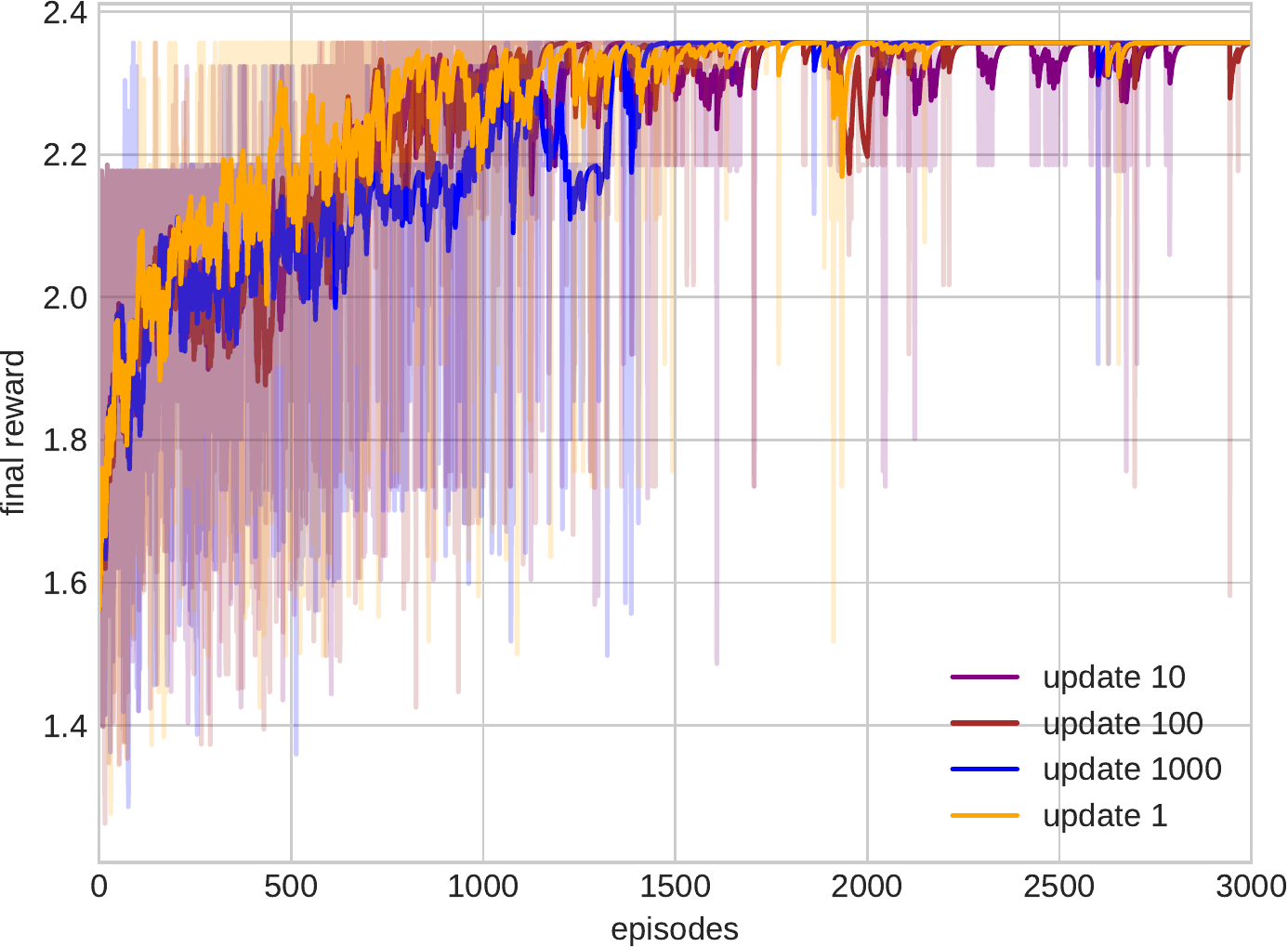}
	\caption{Adjustment of the update frequency}
	\label{fig16}
\end{figure}

The update frequency of the double network parameters, $n_{update}$, is set to 1, 10, 100, and 1000, as shown in Fig.\ref{fig16}. When the update interval is too long, as shown in the curve with $n_{update}$ equal to 100 in the figure, the agent converges slowly and fluctuates greatly. The update frequency, as shown in the curve with $n_{update}$ equal to 1, cause frequent small fluctuations in the curve. The former is because the update is too slow for the agent to learn new things in the target network, while the latter is due to the loss value not smoothly decreasing caused by frequent updates.

\subsection{Prformance analysis}\label{sec6.2}
After training the DHRL-FNMR agent with 12 Network State Information (NLI), it was compared with the KMB algorithm and DRL-M4MR algorithm. The KMB algorithm was compared with three different weights, residual bandwidth, link delay, and packet loss, represented as $KMB_{bw}$, $KMB_{delay}$, and $KMB_{loss}$, respectively. The comparison results are shown below.

\begin{figure}[h]
	\centering
	\includegraphics[width=0.9\linewidth]{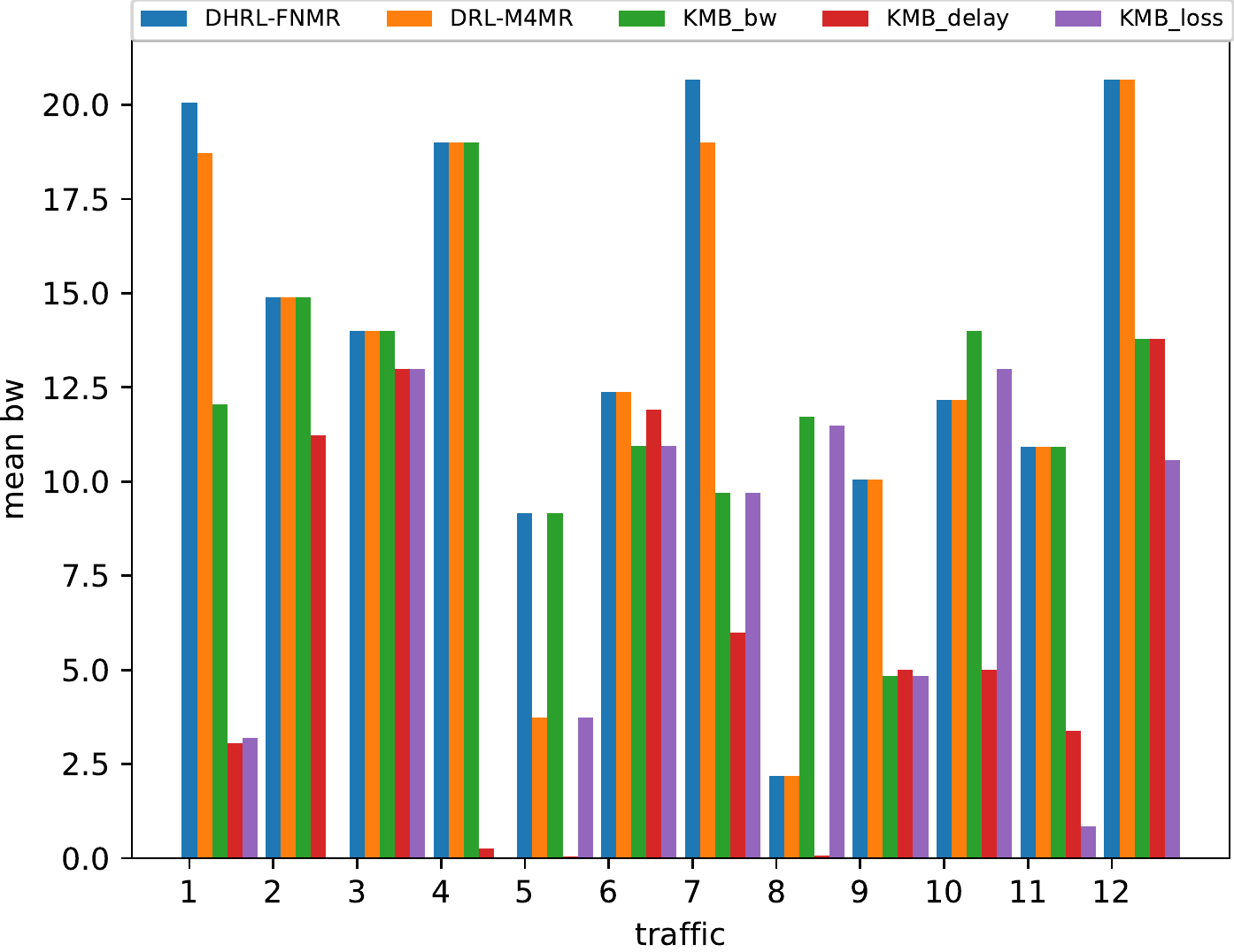}
	\caption{Comparison of average bottleneck bandwidth}
	\label{fig17}
\end{figure}

The metric measured in Fig.\ref{fig17} is the average minimum bottleneck bandwidth of paths from the source node to all destination nodes in the multicast tree, which represents the average bottleneck bandwidth. The DHRL-FNMR agent in this study constructs multicast trees with an average bandwidth 21.34\% higher than that of the $KMB_{bw}$ algorithm, with the highest being 113.09\% higher and the worst being 81.42\% lower than the $KMB_{bw}$ algorithm. There were no cases in which the average bottleneck bandwidth of constructing multicast trees was 0 for $KMB_{delay}$ and $KMB_{loss}$. The DHRL-FNMR algorithm constructs multicast trees with an average bottleneck bandwidth 13.42\% higher than that of the DRL-M4MR algorithm, with the highest being 145.1\% higher. In most scenarios, the average bottleneck bandwidth of the multicast trees constructed by DHRL-FNMR and DRL-M4MR algorithms are the same. However, in the 8th and 10th NLI, the average bottleneck bandwidth is lower than that of $KMB_{bw}$, which is due to the phenomenon of knowledge forgetting. When the agent learns under different NLIs, it can be seen as learning multiple different tasks, and changing the relevant weights during the training process can damage previous actions, leading to knowledge forgetting. This result indicates that the DHRL-FNMR agent can construct multicast trees with larger bottleneck bandwidths.

\begin{figure}[h]
	\centering
	\includegraphics[width=0.9\linewidth]{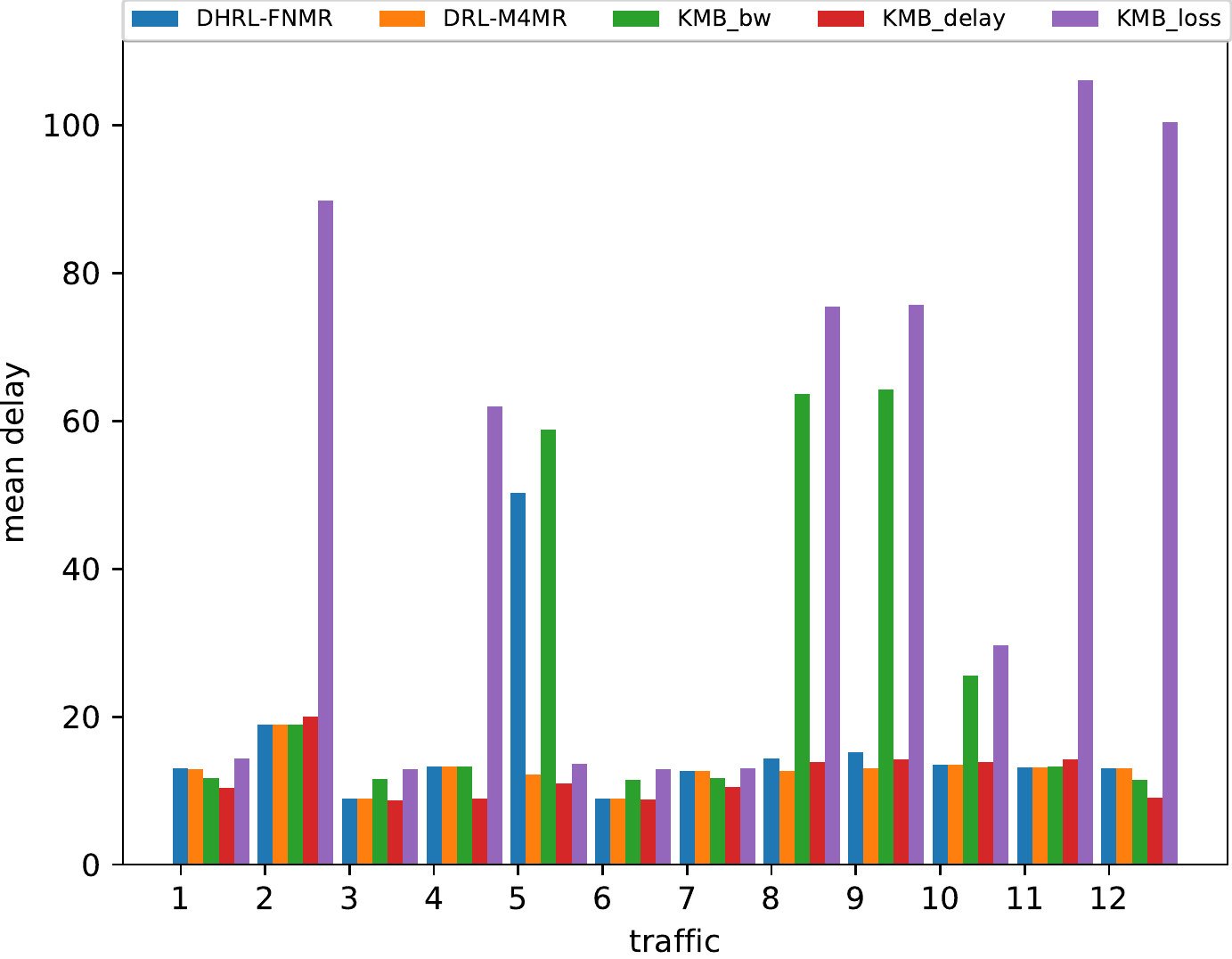}
	\caption{Comparison of average delay}
	\label{fig18}
\end{figure}

Fig.\ref{fig18} measures the average delay of the multicast tree. The average delay of the multicast tree constructed by the HDRL-FNMR agent is 19.06\% and 29.28\% lower than that of the $KMB_{bw}$ and $KMB_{loss}$ algorithms, respectively, and 41.25\% higher than that of the $KMB_{delay}$ algorithm. The best result is 77.51\%, 7.43\%, and 87.62\% lower than that of the $KMB_{bw}$, $KMB_{delay}$, and $KMB_{loss}$ algorithms, respectively. The average delay of the multicast tree constructed by the HDRL-FNMR algorithm is 28.57\% higher than that of the DRL-M4MR algorithm. For the multicast trees constructed under the 8th and 9th NLI, the HDRL-FNMR algorithm is slightly higher than the DRL-M4MR algorithm, while the average delay in other cases is the same. These results indicate that the HDRL-FNMR agent can construct multicast trees with good delay performance.

\begin{figure}[h]
	\centering
	\includegraphics[width=0.9\linewidth]{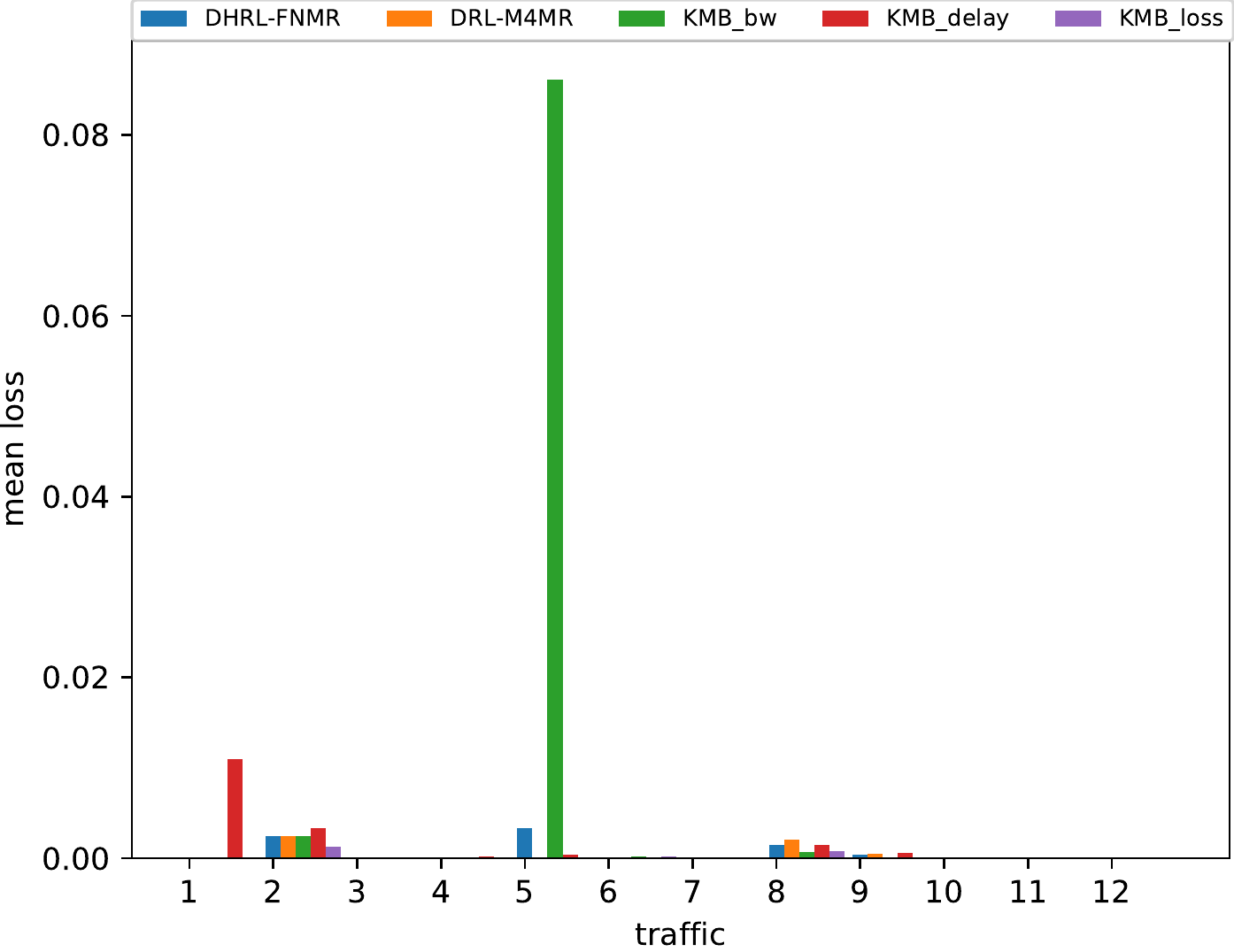}
	\caption{Comparison of average packet loss rate}
	\label{fig19}
\end{figure}

Fig.\ref{fig19} shows the average packet loss rate of multicast trees. The multicast tree constructed by the HDRL-FNMR agent has an average packet loss rate that is 0.22\% lower than that of the multicast tree constructed by the $KMB_{bw}$ algorithm, but is 40.36\% and 13.92\% higher than those constructed by the $KMB_{delay}$ and $KMB_{loss}$ algorithms, respectively. The HDRL-FNMR algorithm has an average packet loss rate that is 28.57\% higher than the DRL-M4MR algorithm, but it can construct multicast trees with zero average delay and low overall delay, indicating that the HDRL-FNMR algorithm can construct multicast trees with good packet loss performance.

\begin{figure}[h]
	\centering
	\includegraphics[width=0.9\linewidth]{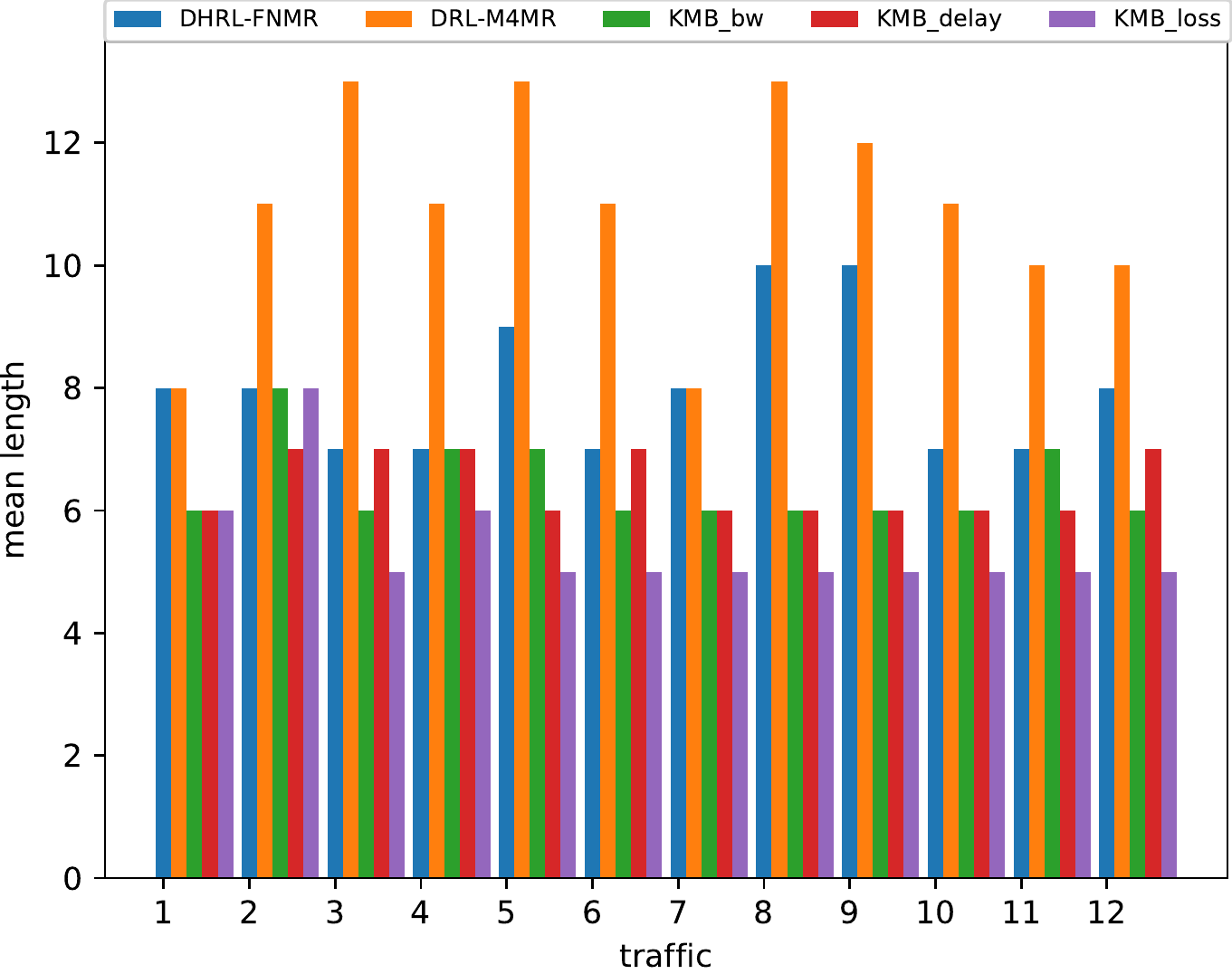}
	\caption{Comparison of average length}
	\label{fig20}
\end{figure}

Fig.\ref{fig20} measures the length of multicast trees. The multicast tree constructed by the HDRL-FNMR algorithm has an average length that is 13.69\%, 13.49\%, and 36.11\% longer than those constructed by the KMB algorithms, respectively. In the worst case, it is 33.33\%, 33.33\%, and 60.0\% longer than the KMB algorithms, respectively. The DRL-M4MR algorithm has redundant branches, so the average length of the multicast tree constructed by the HDRL-FNMR algorithm is 25.25\% lower than that of the DRL-M4MR algorithm, indicating that the HDRL-FNMR algorithm can eliminate the problem of redundant branches caused by the DRL-M4MR algorithm.

\begin{figure}[h]
	\centering
	\includegraphics[width=0.8\linewidth]{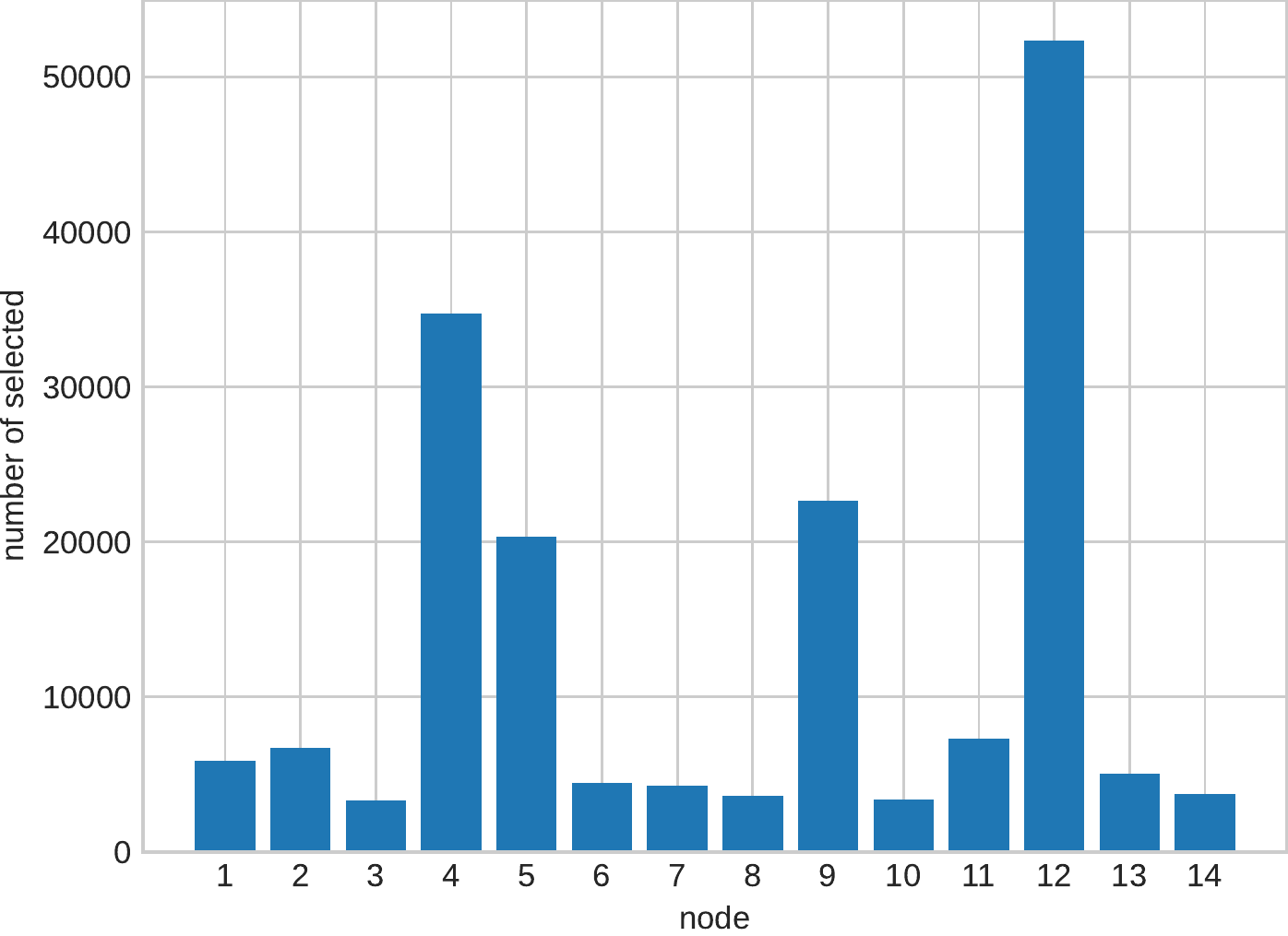}
	\caption{The number of times a node is selected as a subgoal}
	\label{fig21}
\end{figure}

Fig.\ref{fig21} shows the number of times the upper-level meta-controller in the HDRL-FNMR agent selects different nodes as subgoals during training. At the beginning of training, the meta-controller explores extensively, attempting to select each node as a subgoal. In the later stages of training, the meta-controller converges based on the reward function and Decay Epsilon-Greedy, selecting actions with higher Q-values as subgoals. Therefore, the nodes $N_4$ and $N_{12}$ are selected more often as subgoals.

\begin{figure}[h]
	\centering
	\includegraphics[width=0.8\linewidth]{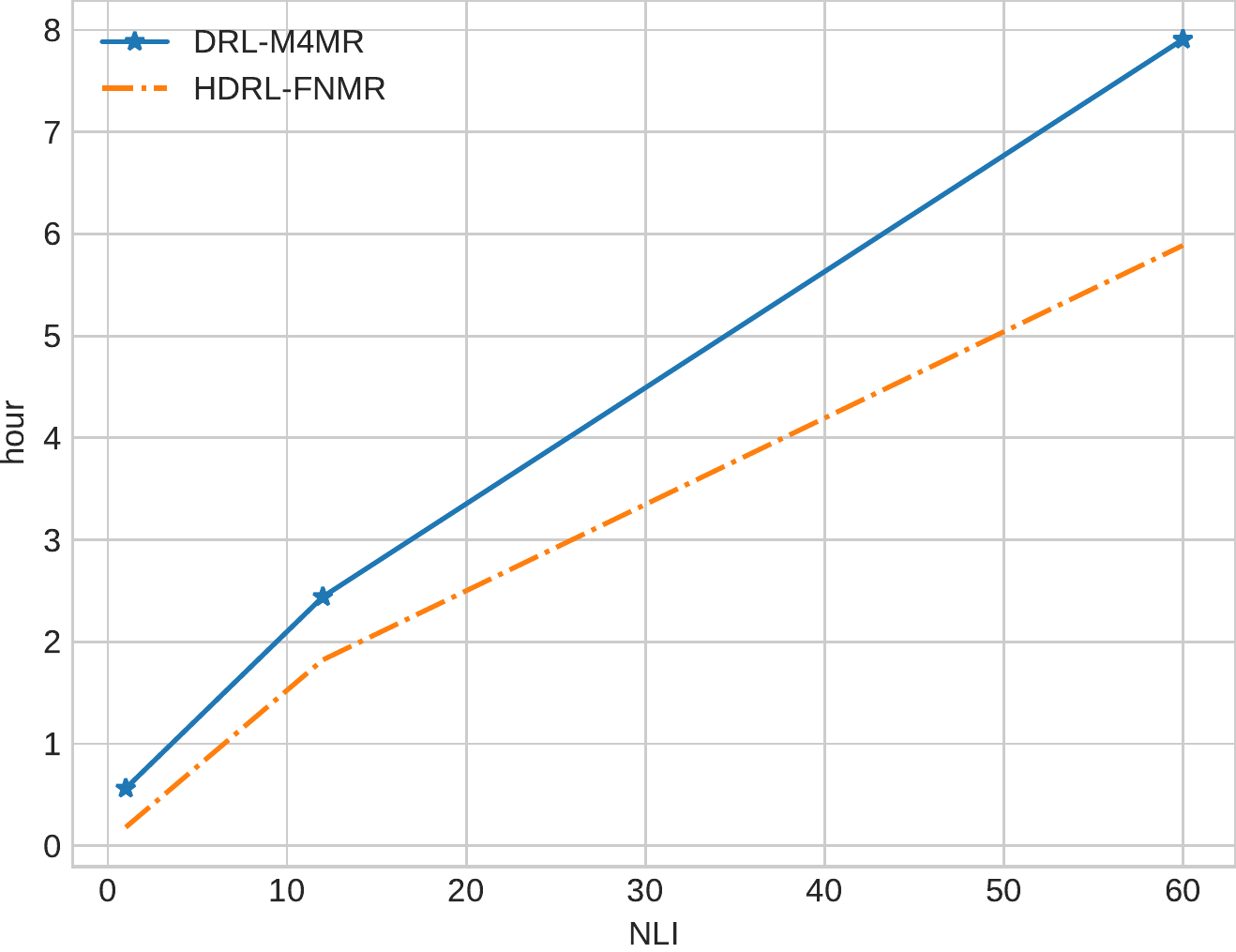}
	\caption{Comparison of training time cost between DRL-M4MR and HDRL-FNMR}
	\label{fig22}
\end{figure}

With the total number of training iterations set to 4000 and the convergence iterations of Decay $\varepsilon$-greedy set to 700, the average training time with 12 NLIs for DRL-M4MR is 2 hours and 26 minutes, while DHRL-FNMR only takes an average of 1 hour and 48 minutes. The comparison of time overhead is shown in Fig.\ref{fig22}. The action space design of DHRL-FNMR algorithm reduces the negative samples of repeatedly selecting invalid actions in the initial training stage, effectively improving search efficiency and significantly reducing learning time for the agent.

\section{Conclusion}\label{sec7}

We introduce the DHRL-FNMR algorithm, which leverages a hierarchical reinforcement learning approach to solve the multicast problem. The problem is divided into two sub-problems: the upper layer is responsible for selecting fork nodes, whereas the lower layer constructs the point-to-point path from the fork node to a destination. We improve upon the DRL-M4MR algorithm by adding the sub-goal matrix to the state space of the intrinsic controller. We also design the action spaces for both intrinsic and extrinsic controllers, and propose a method for finding the fork node and constructing a multicast tree to the destination node. The different actions executed in the action space are categorized and analyzed based on various contexts. Different reward functions are designed for different action situations, and we continue with the design of single-step and final rewards from the DRL-M4MR algorithm. To prevent the overestimation problem during training, we use a double network. We also use prioritized experience replay to balance exploration and exploitation of the agent.

The SDN enables the agent to collect network link information and construct a multicast tree that has improved bandwidth, decreased delay, and lower packet loss rate. The DHRL-FNMR algorithm produces a multicast tree without redundant branches compared to the DRL-M4MR algorithm. Moreover, the multicast tree generated using DHRL-FNMR algorithm outperforms KMB and DRL-M4MR algorithms, in terms of average bandwidth, average delay, and average packet loss rate. The agent's search efficiency is also higher when compared to other algorithms. The experimental results demonstrate the ability of the DHRL-FNMR agent to intelligently adjust the optimal multicast path while flexibly forwarding multicast data in response to changes in network parameters.

\bmhead{Acknowledgments}

This work was supported in part by the National Natural Science Foundation of China (Nos. 62161006, 62172095), the Natural Science Foundation of Fujian Province (Nos. 2020J01875 and 2022J01644), the subsidization of the Innovation Project of Guangxi Graduate Education (Nos. YCSW2022270, YCSW2023310), and Guangxi Key Laboratory of Wireless Wideband Communication and Signal Processing (Nos. GXKL06220110, CRKL220206).

\bmhead{Conflict of Interest}
The authors declare that they have no known competing financial interests or personal relationships that could have appeared to influence the work reported in this paper.

\bmhead{Data Availability}
Data will be made available on request. Code and DRL model are available at https://github.com/GuetYe/DHRL-FNMR.

\bibliography{sn-bibliography}
\bibliographystyle{sn-basic}

\end{document}